\documentclass[a4paper,twoside]{article}

\usepackage{epsfig}
\usepackage{subcaption}
\usepackage{calc}
\usepackage{amssymb}
\usepackage{amstext}
\usepackage{amsmath}
\usepackage{amsthm}
\usepackage{multicol}
\usepackage{pslatex}
\usepackage{apalike}
\usepackage{algorithm2e}
\usepackage[bottom]{footmisc}
\usepackage{SCITEPRESS}     

\usepackage[table]{xcolor}

\usepackage{adjustbox}
\usepackage{mwe}

\usepackage{makecell}
\usepackage{diagbox}
\usepackage{tikz}
\usepackage{multirow}
\usepackage{hyperref}
\usepackage{booktabs}

\hypersetup{
    colorlinks=true,       
    linkcolor=blue,        
    citecolor=black,       
    filecolor=magenta,     
    urlcolor=blue          
}

\usepackage{stfloats}

\newsavebox{\tempbox}

\usepackage{float}

\begin{document}

\title{Automatic Labelling for Low-Light Pedestrian Detection}

\author{\authorname{Dimitrios Bouzoulas\sup{1}\orcidAuthor{0009-0007-4665-6020}, Eerik Alamikkotervo\sup{1}\orcidAuthor{0000-0000-0000-0000} and Risto Ojala\sup{1}\orcidAuthor{0000-0001-5104-2911}}
\affiliation{\sup{1}Energy and Mechanical Engineering, Aalto University, Espoo, Finland}
\email{dimitrios.bouzoulas@aalto.fi, eerik.alamikkotervo@aalto.fi, risto.j.ojala@aalto.fi}
}

\keywords{Pedestrian Detection, Low-Light Conditions.}

\abstract{
Pedestrian detection in RGB images is a key task in pedestrian safety, as the most common sensor in autonomous vehicles and advanced driver assistance systems is the RGB camera. Low-light pedestrian detection lacks large public datasets and autolabelling pipelines. This research proposes a solution in the form of an automated infrared-RGB pipeline. The pipeline consists of 1) Infrared detection, where a fine-tuned model for infrared pedestrian detection is used 2) Label transfer process from the infrared detections to their RGB counterparts 3) Training object detection models using the generated labels for low-light RGB pedestrian detection. 
The research was performed using the KAIST dataset.
For evaluation, three object detection models, DETR, YOLO, and RCNN, were trained on generated and ground truth labels. 
When compared on previously unseen images, the results showed that the models trained on generated labels outperformed the ones trained on ground-truth in 5 out of 6 cases for the mAP@50 and LAMR metrics, and outperformed ground-truth on mAP@50-95 in all cases. 
Acquired results indicate that the proposed autolabelling pipeline could be used for scalable annotation of low-light datasets for pedestrian detection. The source code for this research is available on GitHub 
(\href{https://github.com/BouzoulasDimitrios/IR-RGB-autoamed-low-light-pedestrian-labelling}{link})
\url{https://github.com/BouzoulasDimitrios/IR-RGB-autoamed-low-light-pedestrian-labelling} 
}


\onecolumn \maketitle \normalsize \setcounter{footnote}{0} \vfill

\section{\uppercase{Introduction}}
\label{sec:introduction}

\begin{figure}[t]
    \begin{flushleft}
    \includegraphics[width=0.5\textwidth]{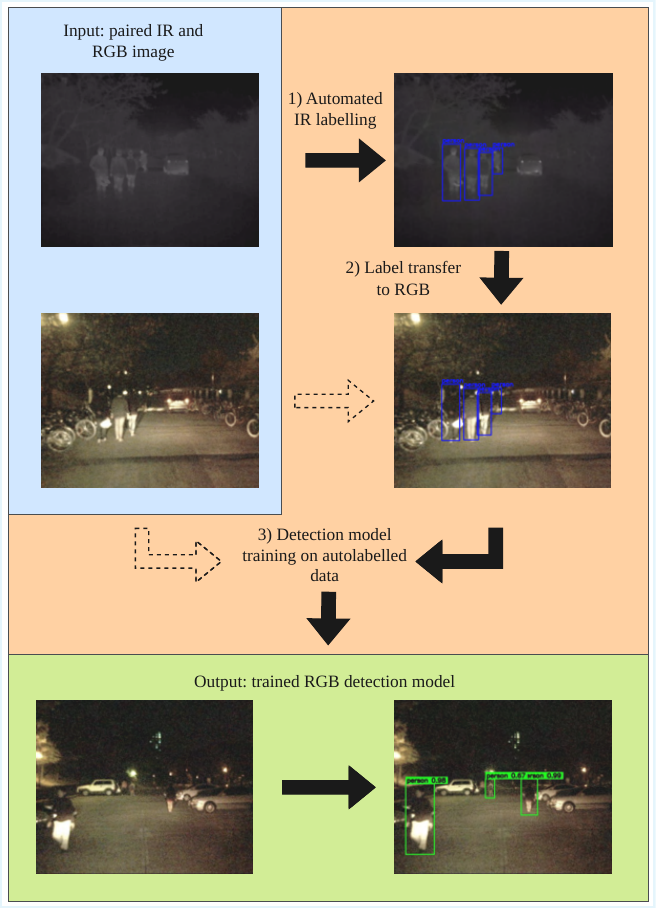}
    \caption{Autolabelling pipeline, using unlabelled low-light IR-RGB image pairs as input, the system outputs labelled RGB images, enabling the training of low-light pedestrian detection models}
    \label{system-diagram}
    \end{flushleft}
\end{figure}

Pedestrian detection has a key role in autonomous vehicles (AVs) and advanced driver assistance systems (ADAS), having the potential to increase safety and lower pedestrian-related traffic accidents \cite{ped_Det}. According to the World Health Organization (WHO) \cite{who2023traffic}, every year approximately 1.19 million people lose their lives in traffic accidents. The crashes referenced by WHO are a result of a multitude of factors, such as poor infrastructure, mechanical problems, and human errors. Looking at the various types of operating conditions, nighttime/low-light conditions pose a common and unavoidable threat. This has been verified by the data analysis done in \cite{moore2024nightdriving} on the data provided by the Transportation Statistics and the National Highway Traffic Safety Administration of the United States, where it was found that the probability of a fatal accident during nighttime was nine times higher compared to daytime. These statistics verify the need for systems capable of reliably detecting pedestrians in low-light conditions.


AVs and cars with ADAS currently on the roads and in production have primarily RGB cameras available. Using only RGB cameras can be challenging in low-light conditions, as RGB model accuracy is highly affected by lighting \cite{obj_det_issues}. A crucial part in developing robust models capable of detecting pedestrians in such conditions is large quantities of labelled data. A large limitation in the labelling process is time.

In this research, we propose an automated Infrared (IR) to RGB labelling pipeline aiming to solve the issues posed by nighttime/low-light conditions, omitting the need for human input. For this goal, we leverage the lighting invariance of IR cameras as visible light has minimal effect on IR radiation. In our approach, we utilise the benefits of IR images to detect pedestrians in IR and then transfer the labels to the corresponding RGB image. Our proposed approach relies on aligned IR and RGB images, which can be conveniently collected with a research vehicle equipped with a suitable sensor setup. The developed pipeline enables automated labelling with no manual human effort, leaving only data collection as manual work. The proposed pipeline has not previously been implemented for autolabelling tasks and shows a range of positive initial results. The autolabels generated using the proposed pipeline achieved the goal of performing on par or outperforming the ground truth labels in many cases, in key metrics such as mAP@50, mAP@50-95, and LAMR. Consequently, the pipeline provides all of the upside gained from autolabelling by eliminating human effort completely while resulting in detection models on par with those trained on manually labelled data.


\section{Related Work}

\subsection{Pedestrian Detection}
Early pedestrian detection approaches utilise hand-crafted features such as Histogram of Oriented Gradients (HOG) \cite{10YEARSOF-peddet}. However, currently, the best performance is achieved by deep neural networks that can learn complex detection mechanisms from the data. Originally, convolutional neural network architectures were proposed but recently, transformer-based architectures have also gained popularity in the pedestrian detection task, DETR \cite{carion2020end} being a popular example. Convolutional architectures are typically further divided into one-stage and two-stage detectors. One-stage detectors predict the bounding box coordinates and related class probabilities in a single forward pass. A common example is the YOLO model family \cite{redmon2016you}. Two-stage detectors first propose regions of interest and then classify and refine the proposals. Common examples are RCNN \cite{girshick2014rich} and its predecessors Fast RCNN \cite{girshick2015fast} and Faster RCNN \cite{rcnn_citation}. To support the development of the pedestrian detection models variety of datasets have been presented over the years, e.g. Caltech Pedestrian Dataset \cite{dollar2009pedestrian}, CityPersons dataset \cite{CityPersons-DATASET}, KITTI \cite{KITTI-DATASET}, and WiderPerson \cite{widerperson_dataset}. 

\subsection{Pedestrian detection in low-light conditions}

In low-light conditions, pedestrian detection from conventional visible-spectrum RGB images is significantly more challenging as the lack of visible light leads to low contrast, changes in feature visibility, and severe noise effects. To improve low-light RGB detection, image enhancement methods which aim to restore the low-light image before feeding it to a standard detector have been proposed \cite{cai2023retinexformer,wei2018deep}. Improved performance can be achieved by optimising the enhancement for the object detection task, specifically by end-to-end training \cite{liu2022image} \cite{guo2021dynamic}. However, superior performance can be achieved by leveraging additional sensors like an IR camera or lidar that provide measurements less affected by the lighting conditions. Lidar-camera fusion for object detection has been extensively researched, and \cite{zhang2022robust,bai2022transfusion} are notable examples, including low-light conditions in the experiments. IR-RGB fusion has also been explored in multiple studies. A popular baseline solution is provided in \cite{KAIST_dataset} and recent work presenting a more advanced fusion strategy in \cite{zhang2024tfdet}. In \cite{liu2021deep} \cite{kruthiventi2017low} teacher-student architecture is proposed where the teacher has access to RGB and IR images, while the students can only rely on RGB images to train a reliable RGB detector for low-light conditions. While most of the available data is from daytime, some datasets addressing the low-light conditions have been published. KAIST \cite{KAIST_dataset}, LLVIP \cite{LLVIP_dataset}, FLIR-ADAS \cite{FLIRADAS_dataset} and CVC-14 \cite{cvc14_dataset} provide RGB and IR images, and NightOwls \cite{nightowls_dataset} only provides RGB images. 

\subsection{Automated labelling}

Detection models perform poorly outside their training domain, and manual annotation is one of the main bottlenecks for training data generation. Thus, research efforts have also focused on ways to reduce or remove the need for manual annotations. A common approach is unsupervised domain adaptation, where the model is adapted from a source domain to a target domain, and annotations are only available in the source domain. 
In \cite{domain_adaptation_autolabelling} IR-RGB pedestrian detection model was adapted from the KAIST dataset domain to the CVC-14 dataset domain in an unsupervised manner with performance close to supervised methods. 
Also in \cite{usupervised_ped_det_ir-thermal_autolabelling}, unsupervised domain adaptation yields comparable results to supervised methods when moving between the CVC-14, KAIST and FLIR ADAS datasets. Another way to reduce the need for labelling is to transfer the labels between sensors. If reliable detections are available in one sensor frame, they can be moved to other sensor frames. In \cite{canton2024automatic}, bounding boxes from a pre-trained RGB detector were transferred to the corresponding IR frames to train IR detection models without the need for manual labelling. The potential of the opposite label transfer method, from IR to RGB, was explored during the creation of the LLVIP \cite{LLVIP_dataset} dataset. The dataset consists of aligned IR-RGB frames, and due to pedestrians being more distinct in IR, the manual labelling was performed in IR, with the subsequent step of transferring the labels to RGB. The groundwork done in LLVIP indicates the potential in labelling IR frames and using those labels for RGB detection tasks. The goal of this research is to explore the idea of fully automating the IR to RGB labelling process by autolabelling data in IR and then transferring the labels to corresponding RGB images. This enables scalable data collection and extensive training of low-light pedestrian detection models for RGB images.

\section{Methods}

This paper presents an automatic pedestrian labelling method for RGB image data, leveraging corresponding IR images (Fig.\ref{system-diagram}). The implementation is aimed at enhancing pedestrian detection in low-light conditions. The proposed architecture is presented in Section \ref{sec:architecture}.   
The KAIST Multispectral Pedestrian Detection Benchmark dataset\cite{KAIST_dataset} used in the paper is described in Section \ref{sec:data}, and the implementation details are presented in Section \ref{sec:evaluation}. 

\subsection{Architecture}
\label{sec:architecture}

First, an RGB-IR image pair is taken as input, and pedestrians are detected from the IR image. Then the detections are transferred from the IR frame to the corresponding RGB frame to act as a label. Finally, an RGB prediction model for low-light conditions is trained with the autolabeled data.

\subsubsection{IR detection}
\label{sec:IR-YOLO}

In low-light conditions, pedestrians are usually easier to detect from IR images compared to RGB images. 
IR images capture the IR radiation emitted from the pedestrians, and the amount of IR radiation emitted by pedestrians is roughly invariant to visible light conditions. 
As a result, an IR model trained on daytime data generalizes well to low-light conditions. Additionally, due to the heating effect of the sun, the environment usually emits more IR radiation during the day than at night. Thus, the pedestrian IR footprint stands out from the background better at night. From these premises, we train an IR pedestrian detection model with daytime IR data and use it to detect pedestrians in unseen IR images from low-light conditions. These IR detections are then used as labels for the corresponding RGB frames

\subsubsection{Label transfer}
\label{sec:label_transfer}

To train an RGB detection model, the labels must be transferred from the IR frame to the RGB frame. Accurate label transfer requires synchronisation and known calibration between the IR and RGB frames. Given a pixel in the IR frame, we need to be able to define its position in the RGB frame. To achieve this, the camera matrices must be known, and the principal points of the IR and RGB cameras must be aligned, or at least one of the cameras must also include depth. 

In the KAIST dataset \cite{KAIST_dataset} used in this paper, the principal points of the RGB and IR cameras are aligned using a beam splitter, and the frames are rectified to have the same focal length, yielding aligned IR and RGB frames. Thus, we can directly transfer the labels from the IR frame to the RGB frame. 

\subsubsection{Training RGB detection model}
\label{sec:training}

The IR detections transferred to the RGB frame are used as autolabels for training a low-light pedestrian detection model for RGB images. 
An IR camera is needed for autolabelling, but the trained prediction model is fully RGB image-based.

\subsection{Data}
\label{sec:data}

The proposed autolabelling method is evaluated with the KAIST dataset. KAIST includes $\sim$95k samples in total, where each sample includes an aligned IR-RGB frame pair and manual ground truth annotation for the cyclist, person, and people classes. The people class is used if individuals cannot be distinguished from a group of people. Preliminary experiments confirmed that training models on all three classes resulted in failure to converge due to excessive classification ambiguity, as the cyclist class contained a low number of instances, while the distinction between a few pedestrians and people appeared inconsistent in many cases. To address this issue, data samples containing people and cyclist labels were excluded, keeping only person labels. Of the remaining images, $\sim$52.7k samples are from daytime (sequences 00, 01, 02, 06, 07, 08)  and $\sim$23.7k samples from low-light conditions (sequences 03, 04, 05, 09, 10, 11). The IR models were trained with all of the daytime sequences except sequence 02, which was retained for testing, resulting in $\sim$51k training images and $\sim$7.3k testing images. The low-light RGB models were trained with all of the low-light sequences except sequence 09, which was retained for testing, resulting in $\sim$26.7k training images and $\sim$3.5k testing images.


\subsection{Evaluation}
\label{sec:evaluation}

We chose to evaluate our pipeline with the YOLOv11 \cite{yolov11_ultralytics_paper}, RCNN \cite{rcnn_citation}, and DETR \cite{RF-DETR_2025_citation} architectures. The specific versions chosen were YOLOv11l, Faster RCNN, and RF-DETR. All three models were trained on the daytime IR frames for 20 epochs with the default hyperparameters. These fine-tuned versions of the models are referred to as IR-YOLO, IR-RCNN, and IR-DETR. For all three models, the models were initialised with COCO \cite{COCO_dataset} pretrained weights. The fine-tuned IR models were used for autolabelling the low-light portion of the dataset. To evaluate how the label confidence threshold affects the final results, two label sets were generated with each model, using 0.25 and 0.5 confidence thresholds. Additionally, the models were also trained with ground truth labels, allowing for performance comparisons between models trained on generated and ground truth labels. 


\begin{table}
    \caption{Number of generated autolabels by each IR pedestrian detection model with different confidence thresholds.}
    \begin{center}
        \begin{tabular}{c c c} 
             \toprule
             Model & Confidence & Label Count \\ [0.5ex] 
             \hline
             IR-YOLO & 0.25 & 26712 \\ 
             IR-YOLO & 0.5 & 16388 \\ 
             IR-DETR & 0.25 & 49280 \\ 
             IR-DETR & 0.5 & 29291 \\ 
             IR-RCNN & 0.25 & 21116 \\ 
             IR-RCNN & 0.5 & 19124 \\ 
             \hline
              Ground Truth & - & 26779 \\ 
             \bottomrule
        \end{tabular}
    \end{center}
    \label{ir_model_label_counts}
\end{table}

Same YOLO, DETR, and RCNN variants were also used for the low-light RGB pedestrian detection for reproducibility purposes. Each of the three models was trained on its IR version labels generated with 0.25 and 0.5 confidence thresholds, and the ground truth labels for comparison. This resulted in a total of 9 low-light RGB pedestrian detection models. Similar to the IR models, each model was trained for 20 epochs. Training was stopped at 20 epochs as validation metrics plateaued, and additional epochs did not result in measurable improvement. For the training, the default hyperparameters were used, with COCO-pretrained weights serving as the starting point.

\section{Results}

\begin{figure}
    \centering
    \includegraphics[width=0.5\textwidth]{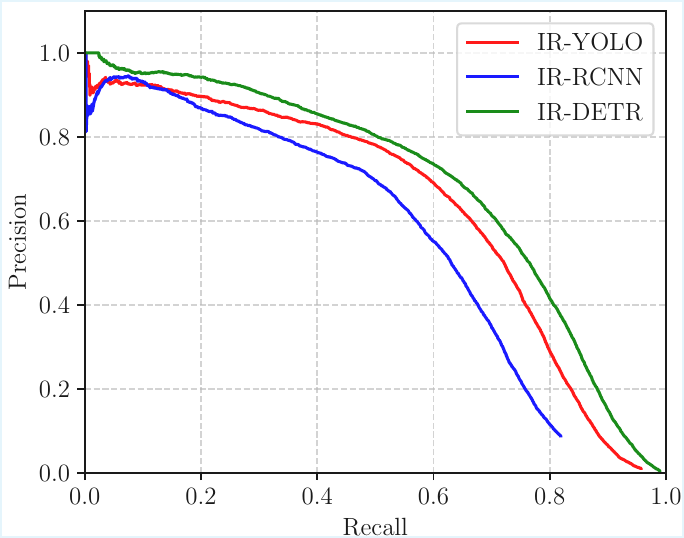}
    \caption{Precision-Recall curves for IR models used for autolabelling.}
    \label{ir_pr_curve}
\end{figure}

The number of labels generated with each of the IR detection models at 0.25 and 0.5 confidence thresholds is presented in Table \ref{ir_model_label_counts}. The KAIST ground truth label count is included for comparison. IR-DETR with 0.25 confidence generated the highest number of labels, and IR-YOLO with 0.5 confidence generated the fewest number of labels. 
To assess the accuracy of the generated labels, standard object detection metrics 
\textbf{Precision(P)}, \textbf{Recall(R)}, \textbf{F1}, \textbf{mAP@50}, \textbf{mAP@50-95} and \textbf{Log Average Miss Rate (LAMR)} are also presented at both confidence levels (Table \ref{ir_model_metrics}). 
IR-DETR had the highest overall performance on the mAP and LAMR metrics, and the labels generated with IR-DETR at 0.5 confidence had the highest F1 score. YOLO had the second-highest and RCNN the lowest performance on the mAP and LAMR metrics. For RCNN and YOLO 0.25 confidence threshold yielded a higher F1 score than a 0.5 confidence threshold. 
The precision-recall curves for the IR models are presented in Figure \ref{ir_pr_curve}.

\begin{table*}
    \centering
    \caption{Autolabelling accuracy metrics for the IR pedestrian detection models with different confidences. Best result \textbf{bolded} in each column.}
    \begin{tabular}{cccccccccc}
        \toprule
         & 
        \multicolumn{3}{c}{\textbf{Conf. 0.5}} &  
        \multicolumn{3}{c}{\textbf{Conf. 0.25}} \\ 
        \cmidrule(lr){2-4} \cmidrule(lr){5-7} 
       \textbf{Model} & \textbf{P} & \textbf{R} & \textbf{F1} & \textbf{P} & \textbf{R} & \textbf{F1} & \textbf{mAP@50} & \textbf{mAP@50-95} & \textbf{LAMR} $\downarrow$ \\
        \midrule 
        IR-YOLO  & \textbf{0.836} & 0.370 & 0.513 & \textbf{0.746} & 0.544 & \textbf{0.630}& 0.635 & 0.245 & 0.564 \\
        IR-RCNN  & 0.768 & 0.391 & 0.519 & 0.743 & 0.435 & 0.549 & 0.554 & 0.208 & 0.623 \\
        IR-DETR  & 0.788 & \textbf{0.528} & \textbf{0.632} & 0.545 & \textbf{0.741} & 0.628 & \textbf{0.688} & \textbf{0.284} & \textbf{0.518} \\
        \bottomrule 
    \end{tabular}
    \label{ir_model_metrics}
\end{table*}

\begin{table*}
    \centering
    \caption{Validation test results comparison with \textcolor{green}{green} for results above the baseline and \textcolor{red}{red} for results below baseline. The precision and recall values have been reported with an IoU threshold of 0.5.}
    
    \setlength{\tabcolsep}{10pt} 
    
    \begin{tabular}{lcccccc} 
    \toprule
    \textbf{Labels} & \textbf{P} & \textbf{R} & \textbf{F1} & \textbf{mAP@50} & \textbf{mAP@50-95} & \textbf{LAMR} $\downarrow$ \\
    \midrule 

    \multicolumn{7}{c}{\textbf{Base Model YOLO}} \\ 
    \midrule 

    IR-YOLO (Conf. 0.5) & \textcolor{green}{0.941} & \textcolor{red}{0.171} & \textcolor{red}{0.289} & \textcolor{green}{0.290} & \textcolor{green}{0.128} & \textcolor{green}{0.733} \\
    IR-YOLO (Conf. 0.25) & \textcolor{green}{0.924} & \textcolor{green}{0.236} & \textcolor{green}{0.376} & \textcolor{green}{0.321} & \textcolor{green}{0.138} & \textcolor{green}{0.707} \\
    KAIST/Ground Truth & 0.902 & 0.196 & 0.322 & 0.261 & 0.116 & 0.758 \\\midrule
    
    \multicolumn{7}{c}{\textbf{Base Model RCNN}} \\ 
    \midrule 

    IR-RCNN (Conf. 0.5) & \textcolor{red}{0.955} & \textcolor{red}{0.202} & \textcolor{red}{0.333} & \textcolor{red}{0.290} & \textcolor{green}{0.125} & \textcolor{red}{0.724} \\
    IR-RCNN (Conf. 0.25) & \textcolor{red}{0.944} & \textcolor{green}{0.221} & \textcolor{green}{0.358} & \textcolor{green}{0.307} & \textcolor{green}{0.127} & \textcolor{green}{0.711} \\
    KAIST/Ground Truth & 0.963 & 0.207 & 0.341 & 0.292 & 0.115 & 0.719 \\\midrule
 
    \multicolumn{7}{c}{\textbf{Base Model DETR}} \\ 
    \midrule 

    IR-DETR (Conf. 0.5) & \textcolor{green}{0.809} & \textcolor{green}{0.337} & \textcolor{green}{0.475} & \textcolor{green}{0.437} & \textcolor{green}{0.178} & \textcolor{green}{0.635}\\
    IR-DETR (Conf. 0.25) & \textcolor{red}{0.722} & \textcolor{green}{0.369} & \textcolor{green}{0.489} & \textcolor{green}{0.442} & \textcolor{green}{0.183} & \textcolor{green}{0.634} \\
    KAIST/Ground Truth & 0.835 & 0.315 & 0.457 & 0.397 & 0.154 & 0.663\\
    \bottomrule
    \end{tabular}
    \label{final_results_color_coded}

\end{table*}

The performance of the low-light RGB models trained with the autolabels is evaluated in Table \ref{final_results_color_coded}. 
The RGB model trained with the manual ground truth annotation acts as the baseline. 
The models trained with the proposed autolabels outperformed the models trained with the ground truth in 5 out of 6 cases in mAP@50 and LAMR metrics, and in 6 out of 6 cases in the mAP@50-95 metric. 
For all models, ground truth label performance in mAP and LAMR metrics was exceeded when using 0.25 label confidence, and for YOLO and DETR, also when using 0.5 label confidence.

\begin{figure}[htbp]
    \centering
\subfloat[YOLO]{%
    \includegraphics[width=0.48\textwidth]{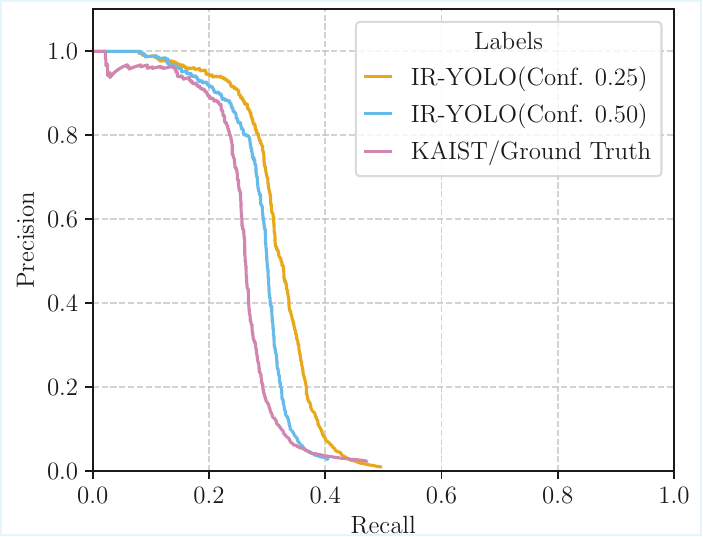}%
    }\\[-0.0em]
    \subfloat[RCNN]{%
        \includegraphics[width=0.48\textwidth]{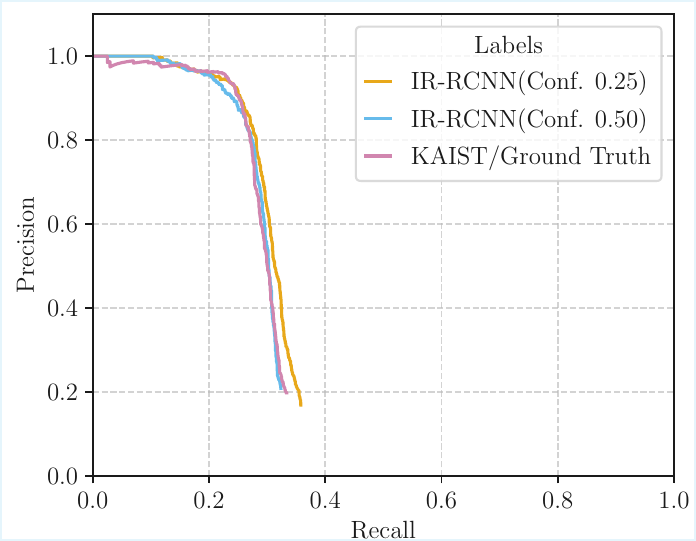}%
    }\\[-0.0em]
    \subfloat[DETR]{%
        \includegraphics[width=0.48\textwidth]{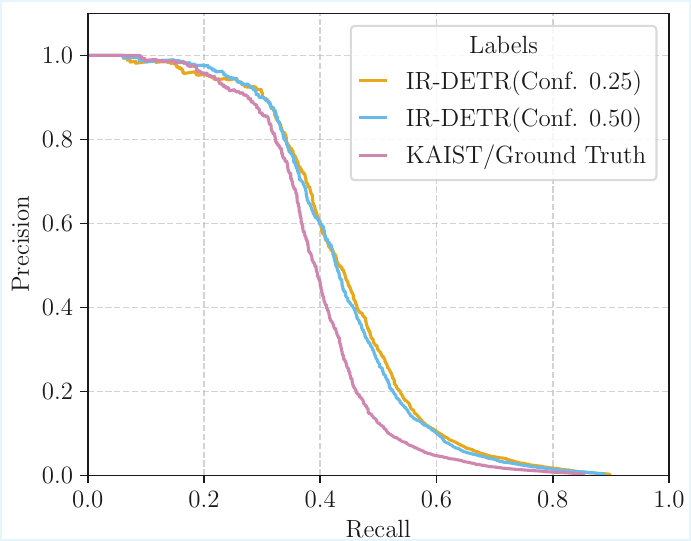}%
    }

    \caption{Precision-Recall (PR) curves for the YOLO, RCNN, and DETR RGB detection models when trained on autolabels generated with the IR models versus ground truth labels.}
    \label{p-r_curves_base_yolov11_combined}
\end{figure}

The best performance overall was achieved by DETR trained on generated labels with a confidence value of 0.25.
The precision-recall curves for each RGB model variant are presented in Figure \ref{p-r_curves_base_yolov11_combined}. Each architecture exhibits quite similar precision-recall tradeoffs regardless of the training labels used, while the differences between architectures are significantly larger. These results showcase the benefit of autolabelling as they performed on par or better than manually generated labels while requiring no human effort or input.

\begin{figure*}
\centering
\resizebox{\textwidth}{!}{ 

\begin{tabular}{c @{\hspace{4pt}} c @{\hspace{4pt}} c @{\hspace{4pt}} c @{\hspace{4pt}} c @{\hspace{4pt}} c @{\hspace{20pt}} c @{\hspace{4pt}} c @{}}

& & \multicolumn{4}{@{\hspace{4pt}}c@{\hspace{4pt}}}{\textbf{Training labels}} & \\[-1.5ex]
& \multicolumn{5}{@{\hspace{4pt}}c@{\hspace{4pt}}}{\rule{\dimexpr 310pt}{0.4pt}} & \\

& & & \bf{IR-DETR 0.25 Conf} & \bf{IR-DETR 0.50 Conf} & \bf{GT labels} & \bf{GT IR} & \bf{GT RGB} \\

\multirow{6}{*}[-150pt]{\rotatebox{90}{\textbf{Prediction model}}} &
\multirow{6}{*}[70pt]{\vrule height 520pt width 0.6pt} &
\multirow{2}{*}[15pt]{\rotatebox{90}{\textbf{DETR}}} &

\includegraphics[width = 1.4in]{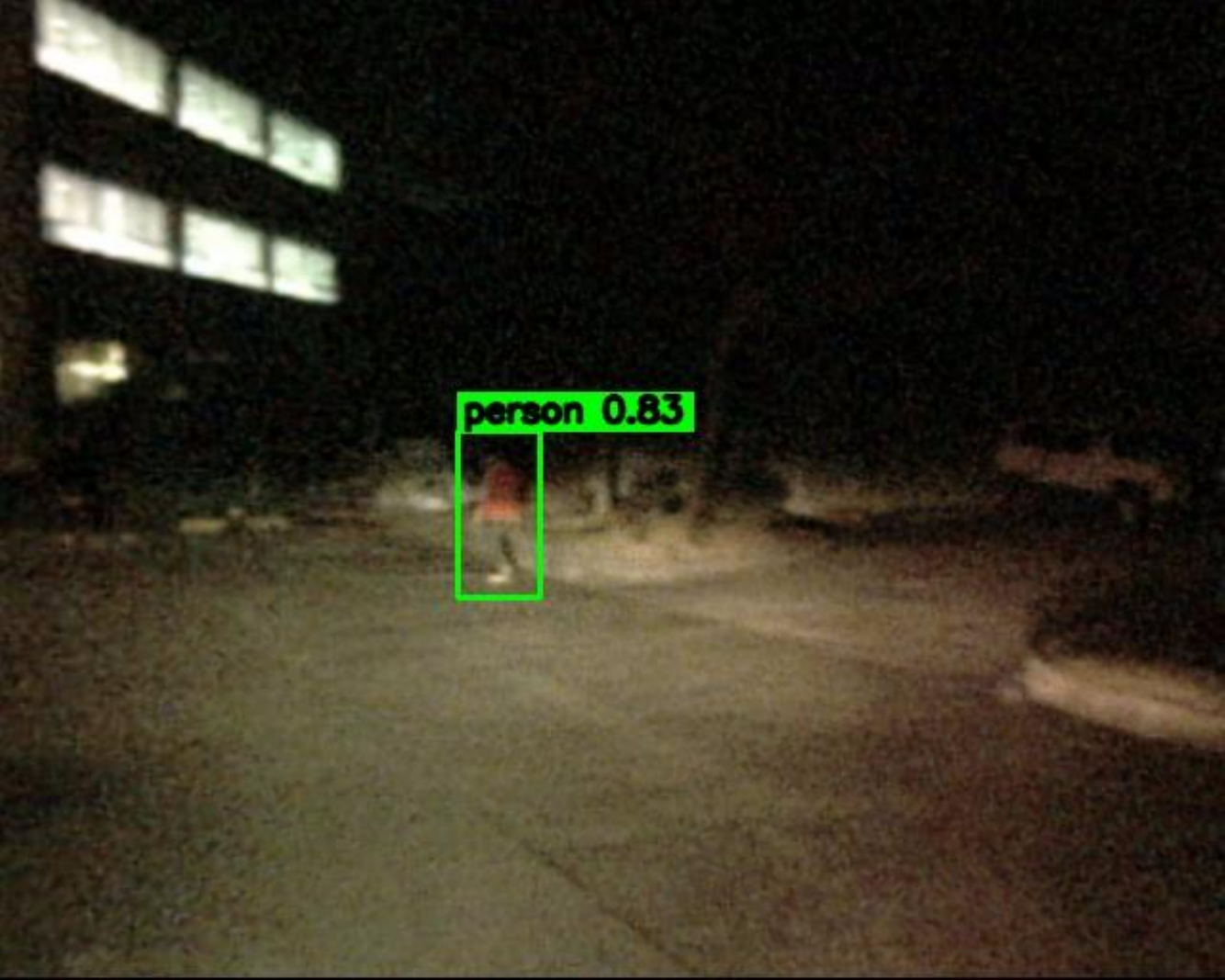} &
\includegraphics[width = 1.4in]{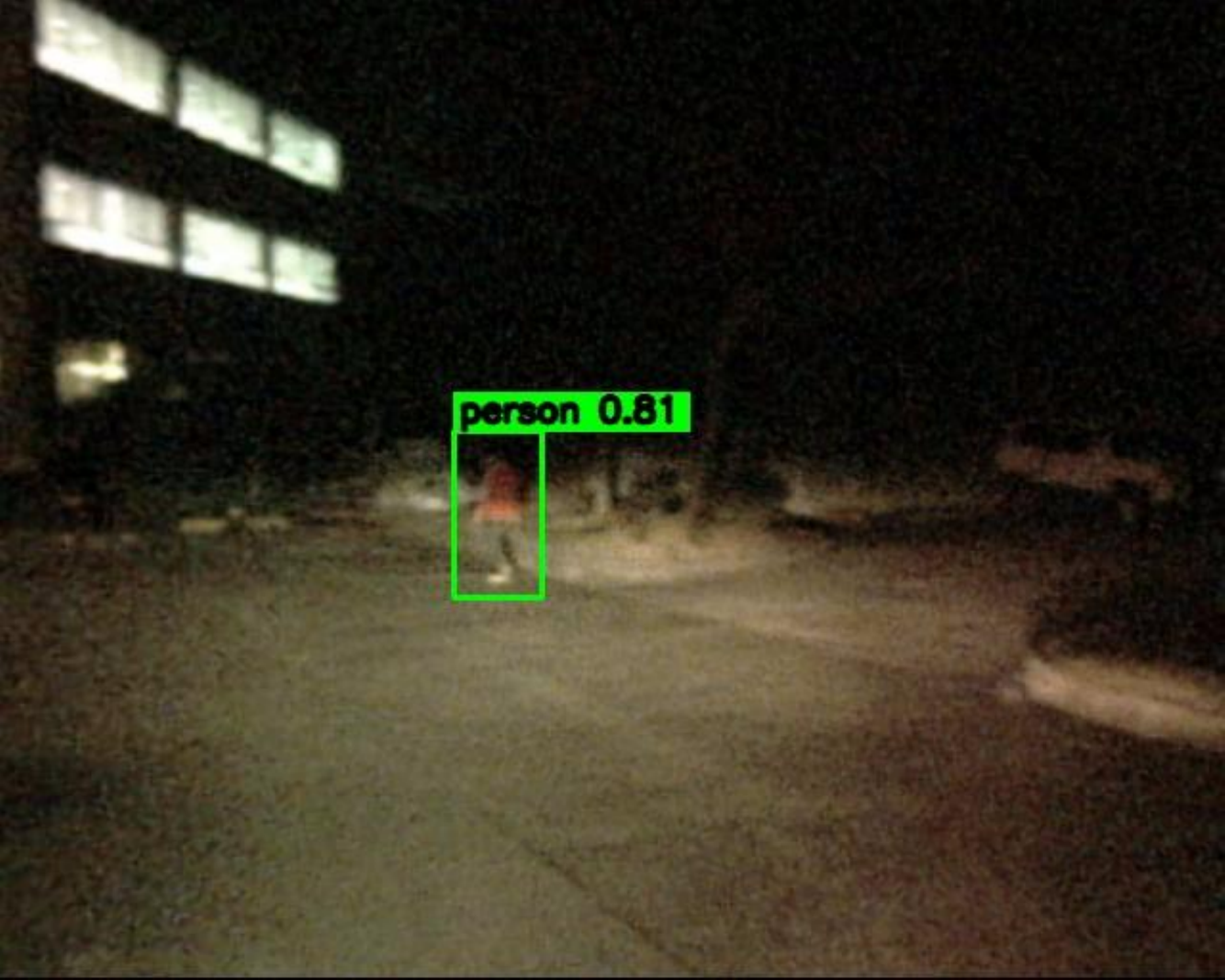} &
\includegraphics[width = 1.4in]{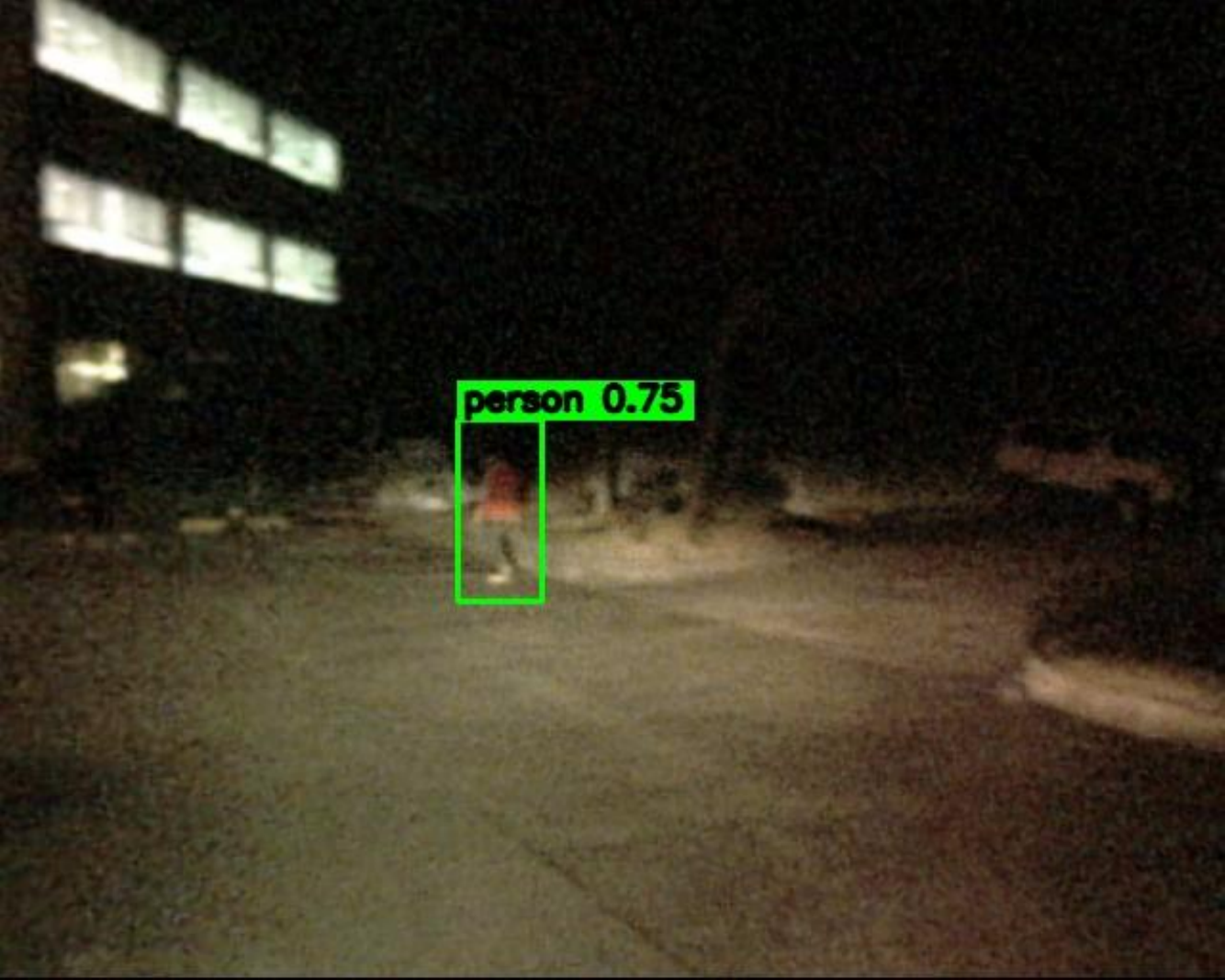} &
\includegraphics[width = 1.4in]{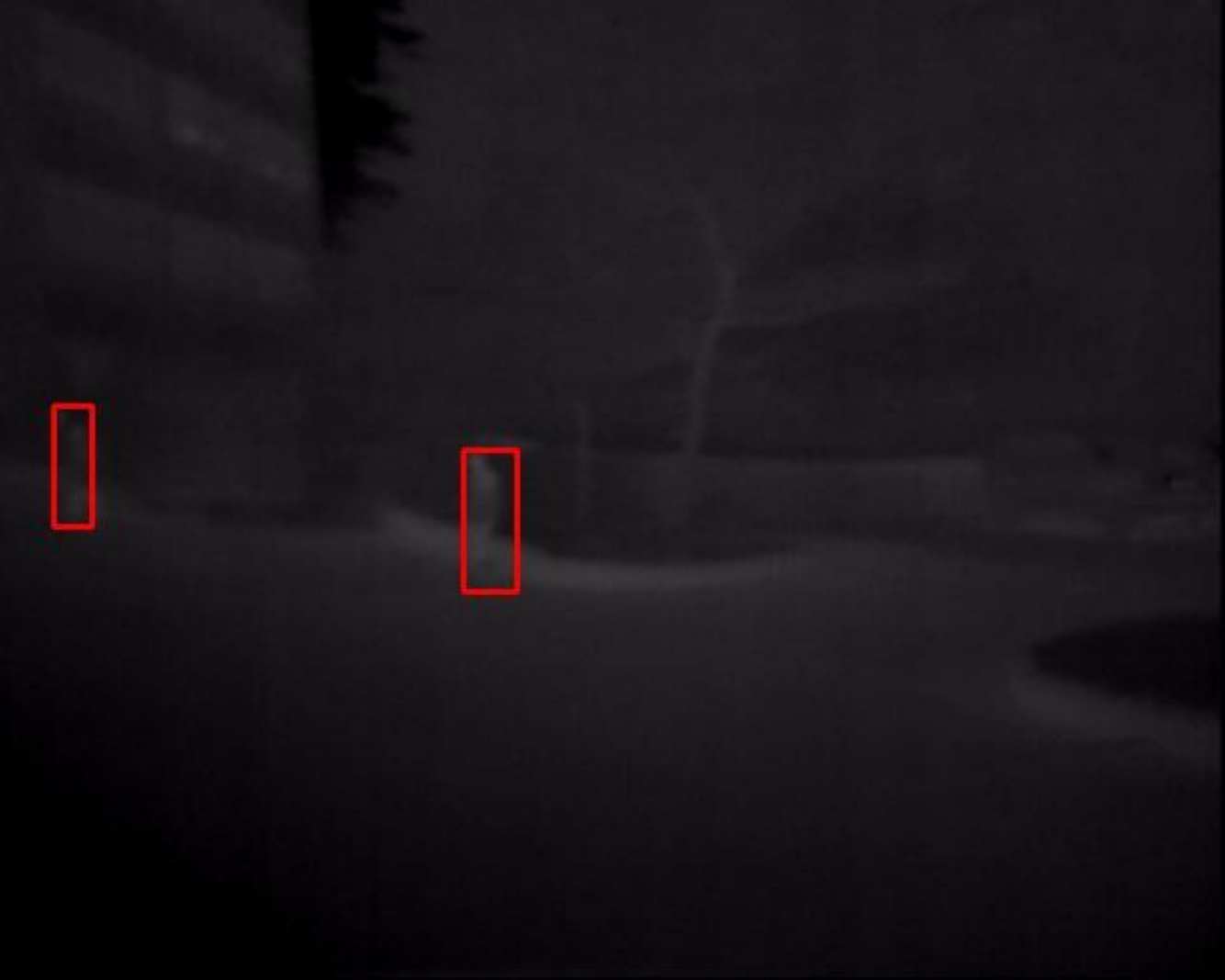} &
\includegraphics[width = 1.4in]{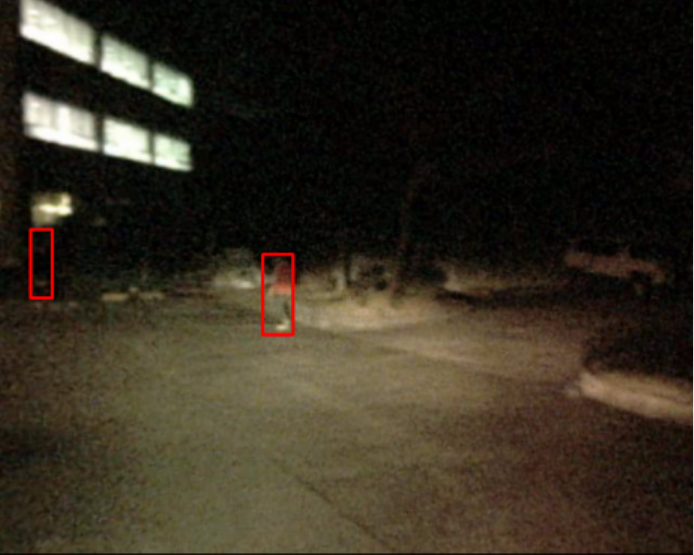} \\

&
&
&
\includegraphics[width = 1.4in]{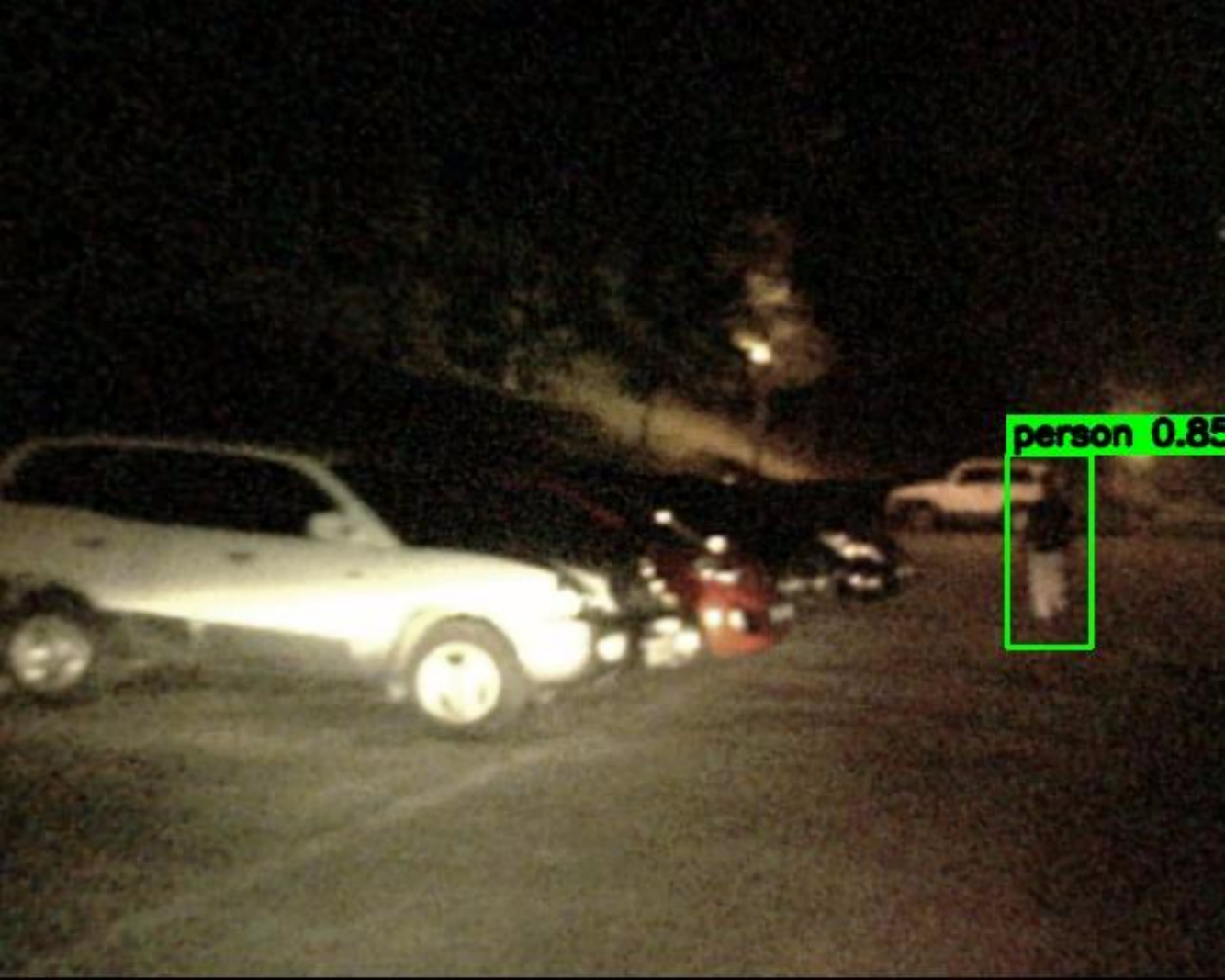} &
\includegraphics[width = 1.4in]{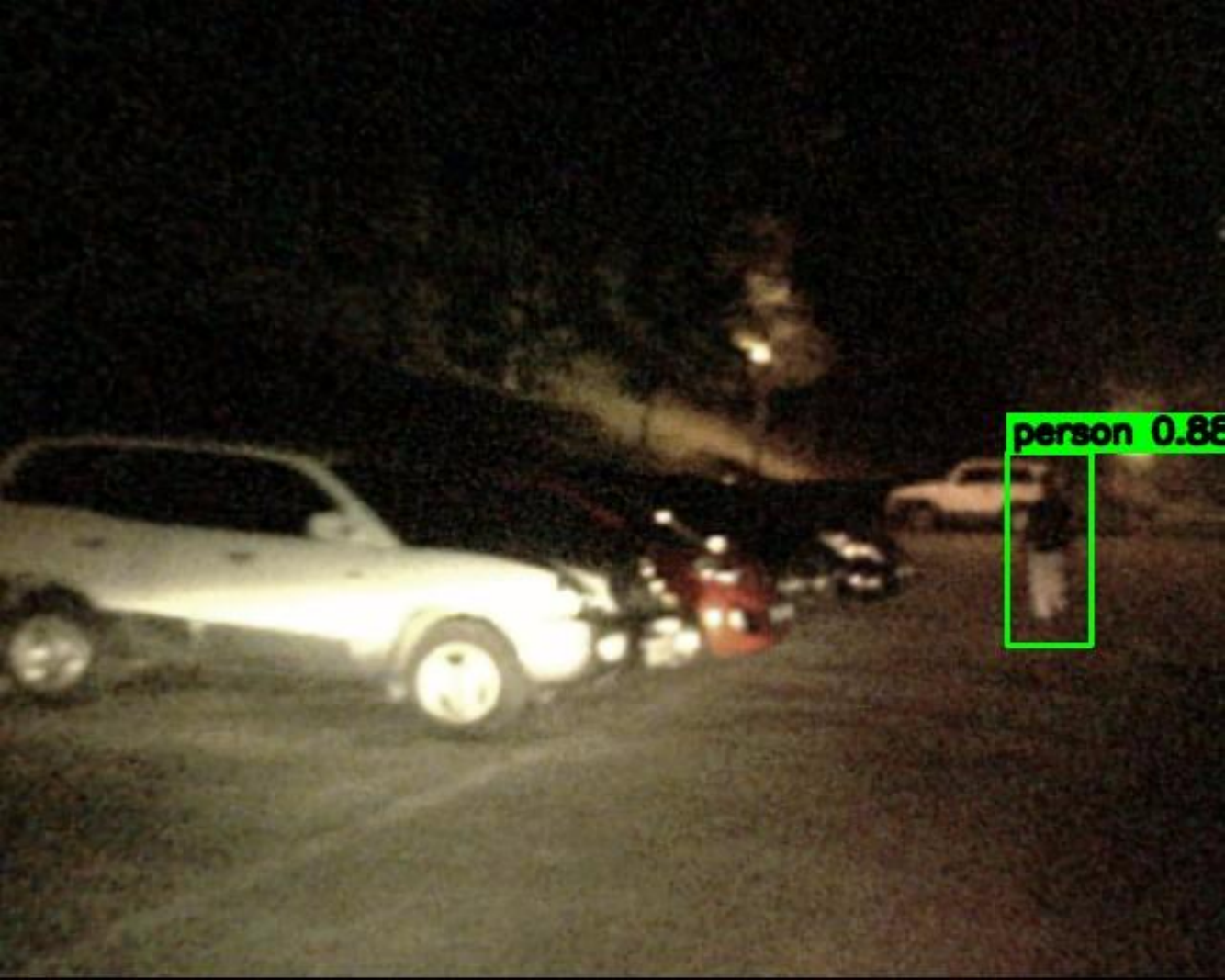} &
\includegraphics[width = 1.4in]{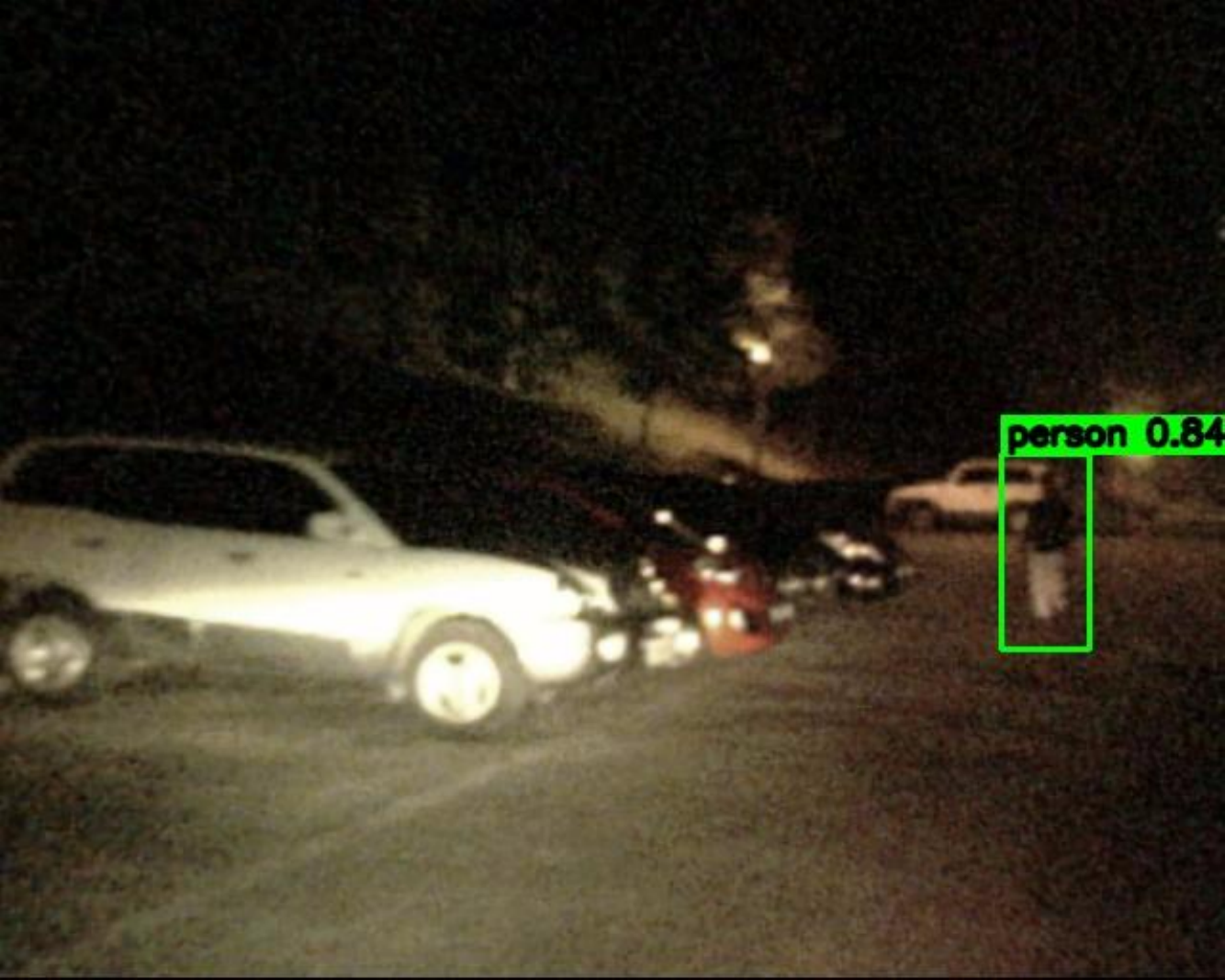} &
\includegraphics[width = 1.4in]{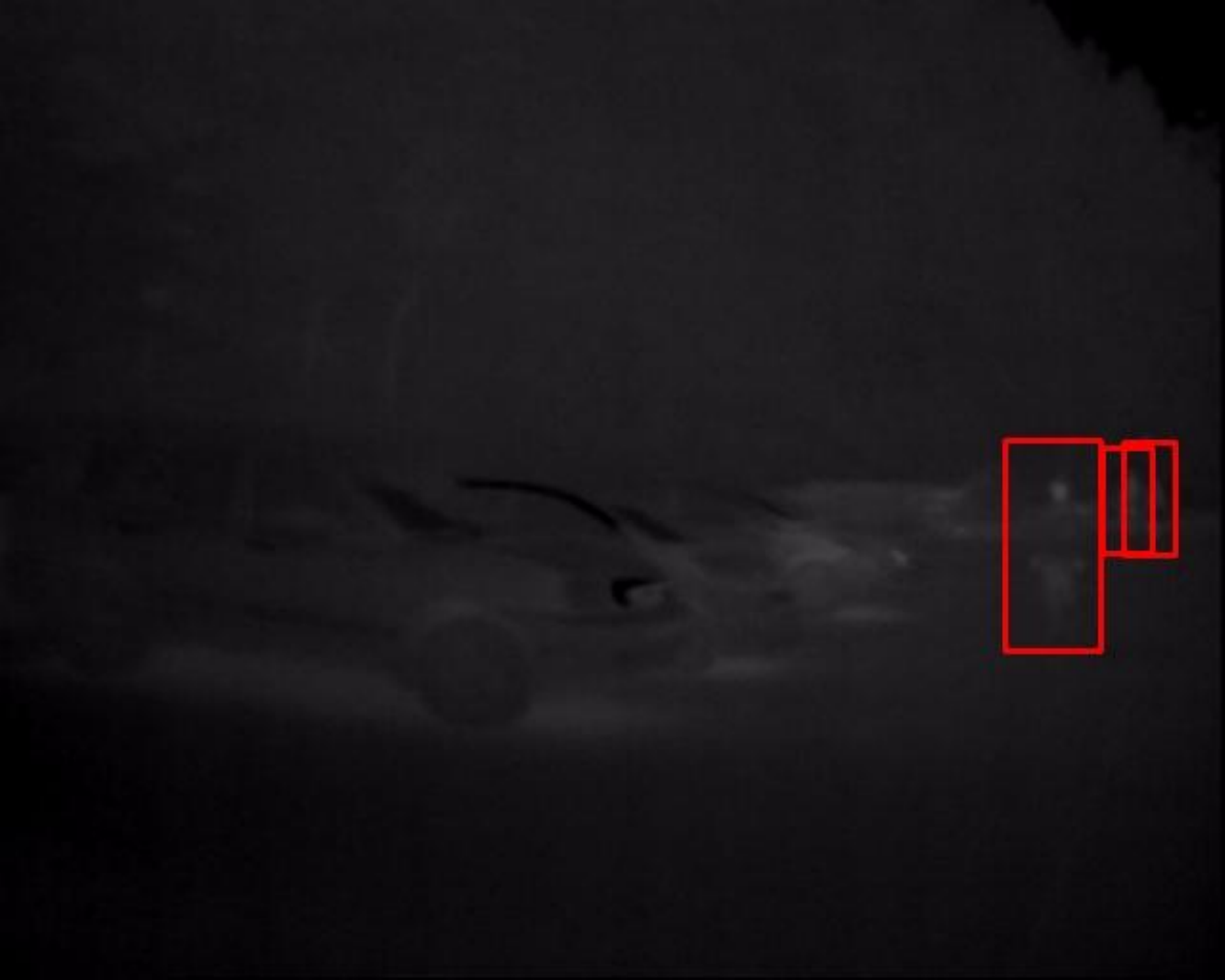} &
\includegraphics[width = 1.4in]{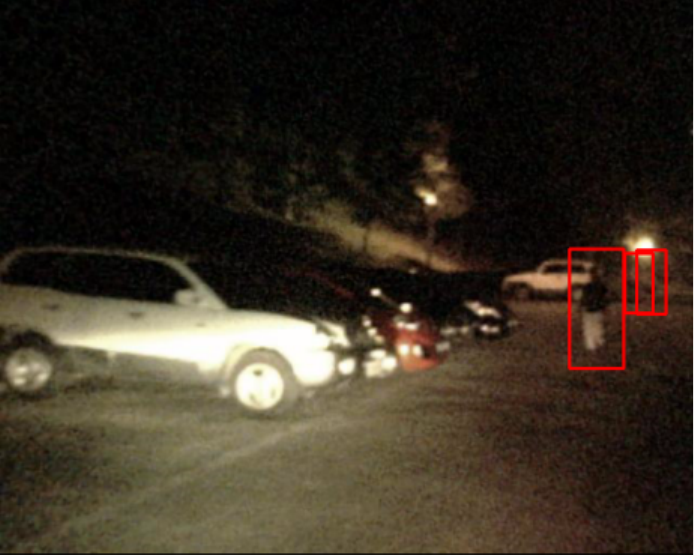} \\
\hspace{1cm} \\

& & & \bf{IR-YOLO 0.25 Conf} & \bf{IR-YOLO 0.50 Conf} & \bf{GT labels} & \bf{GT IR} & \bf{GT RGB} \\
&
&
\multirow{2}{*}[15pt]{\rotatebox{90}{\textbf{YOLO}}} &
\includegraphics[width = 1.4in]{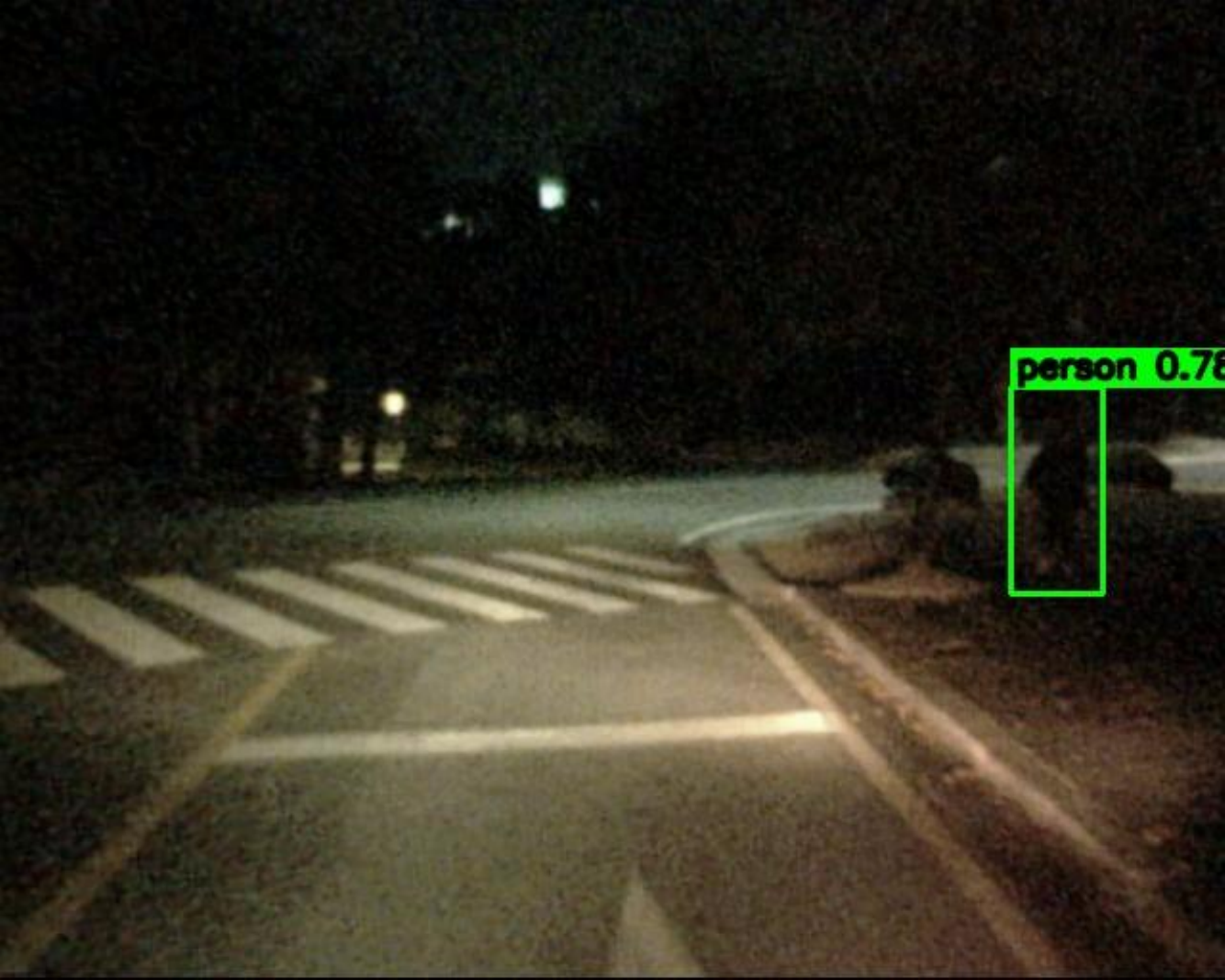} &
\includegraphics[width = 1.4in]{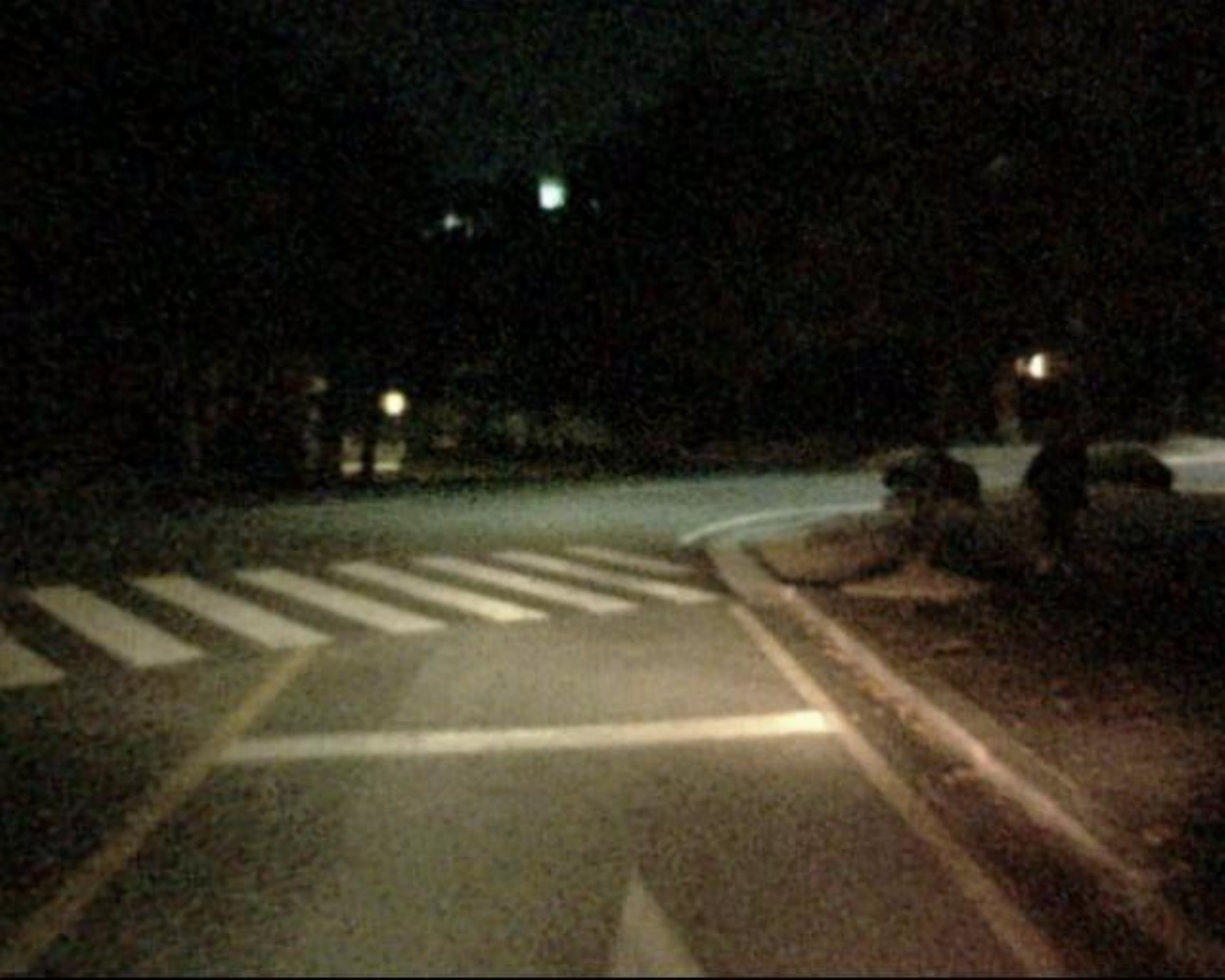} &
\includegraphics[width = 1.4in]{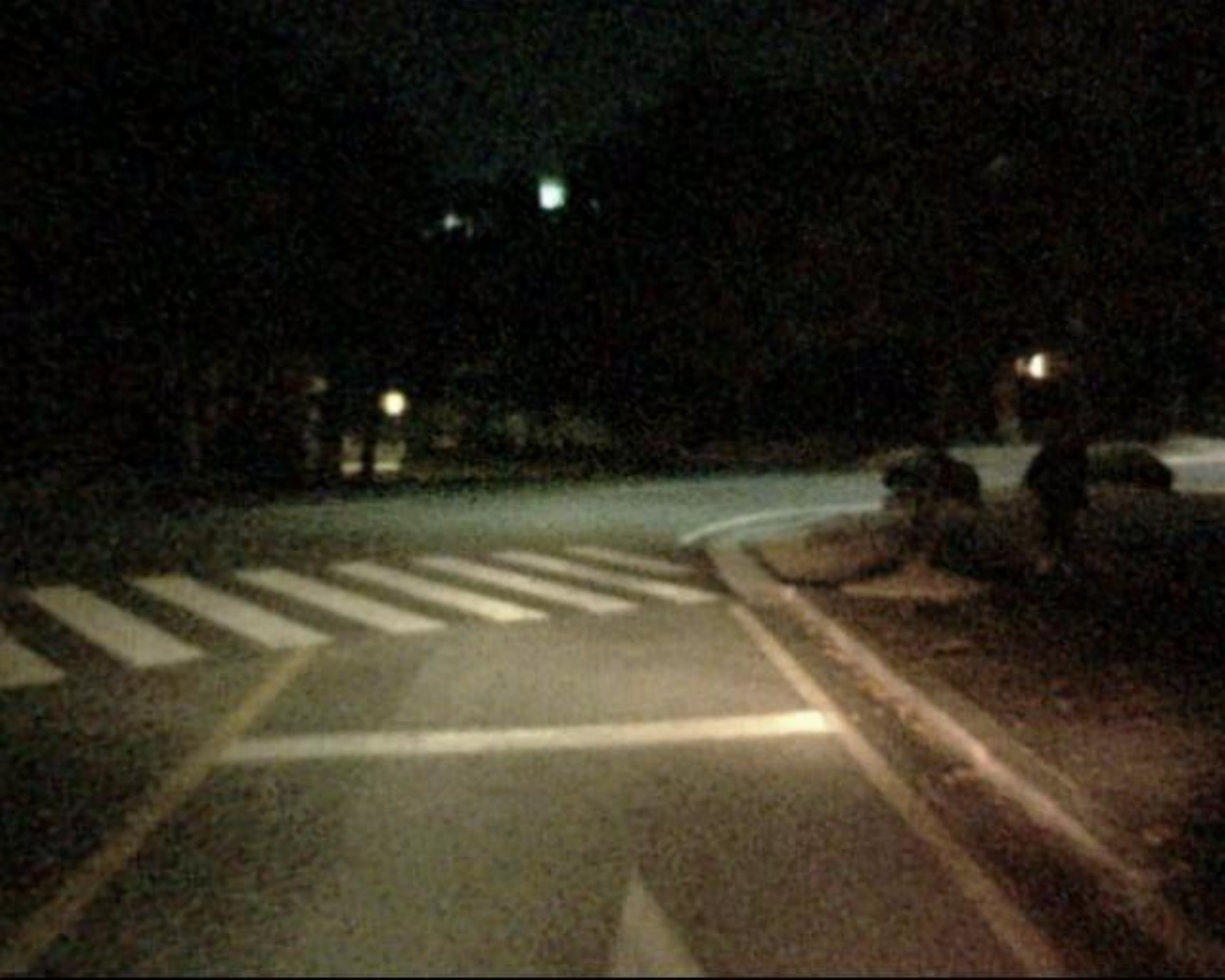} &
\includegraphics[width = 1.4in]{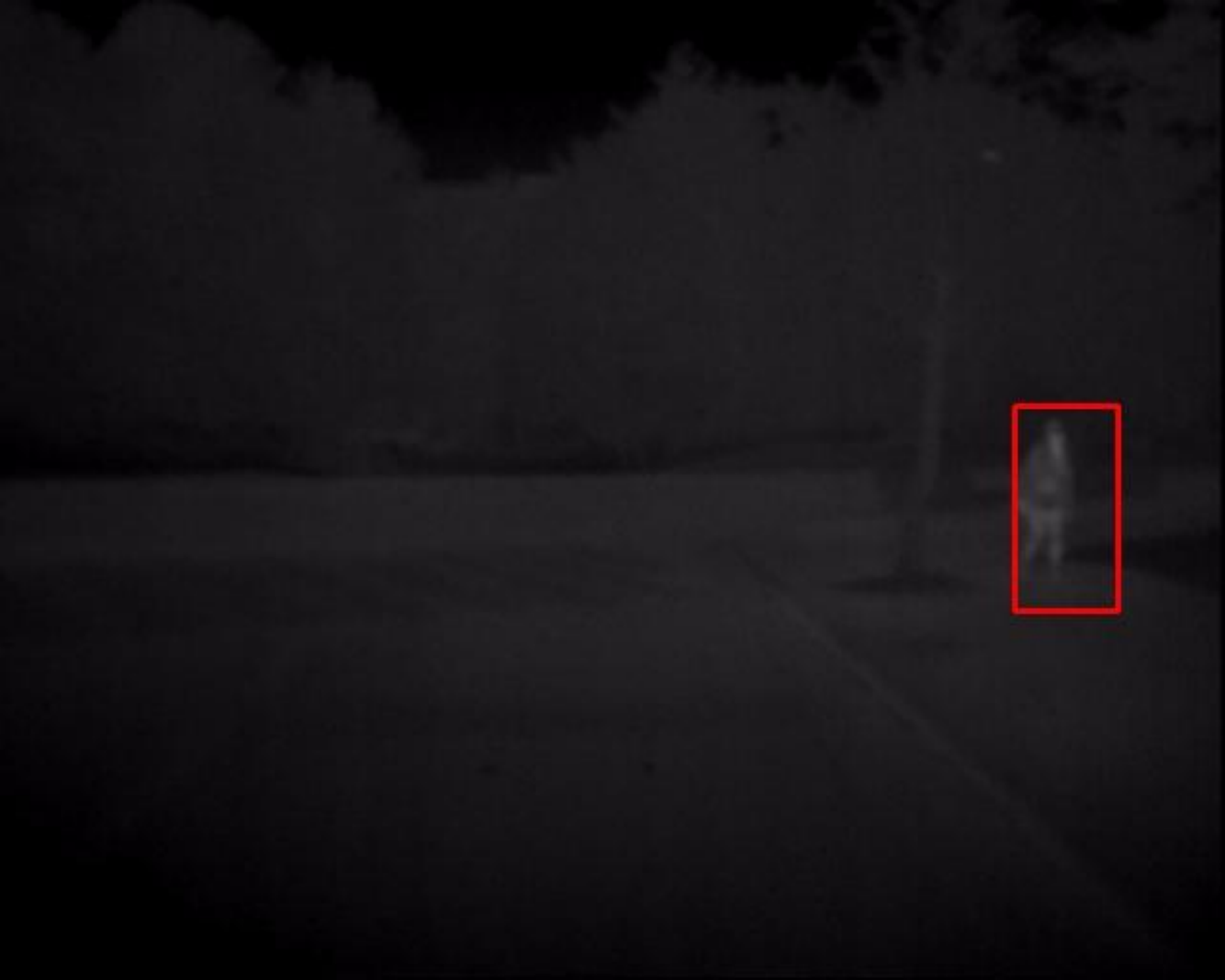} &
\includegraphics[width = 1.4in]{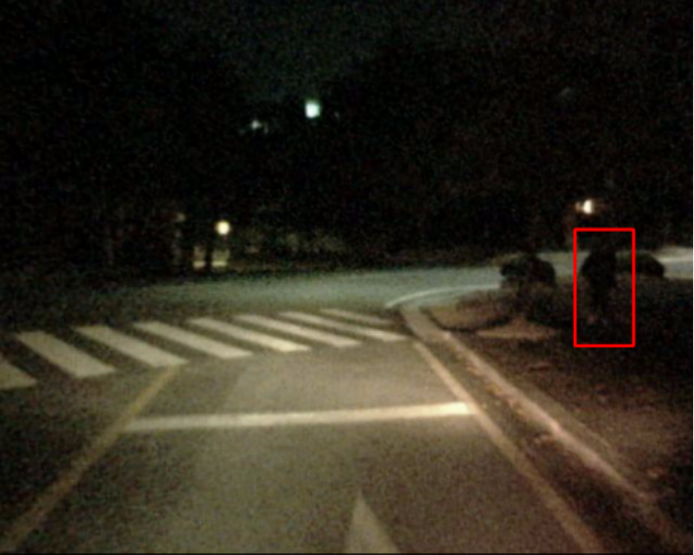} \\

&
&
&
\includegraphics[width = 1.4in]{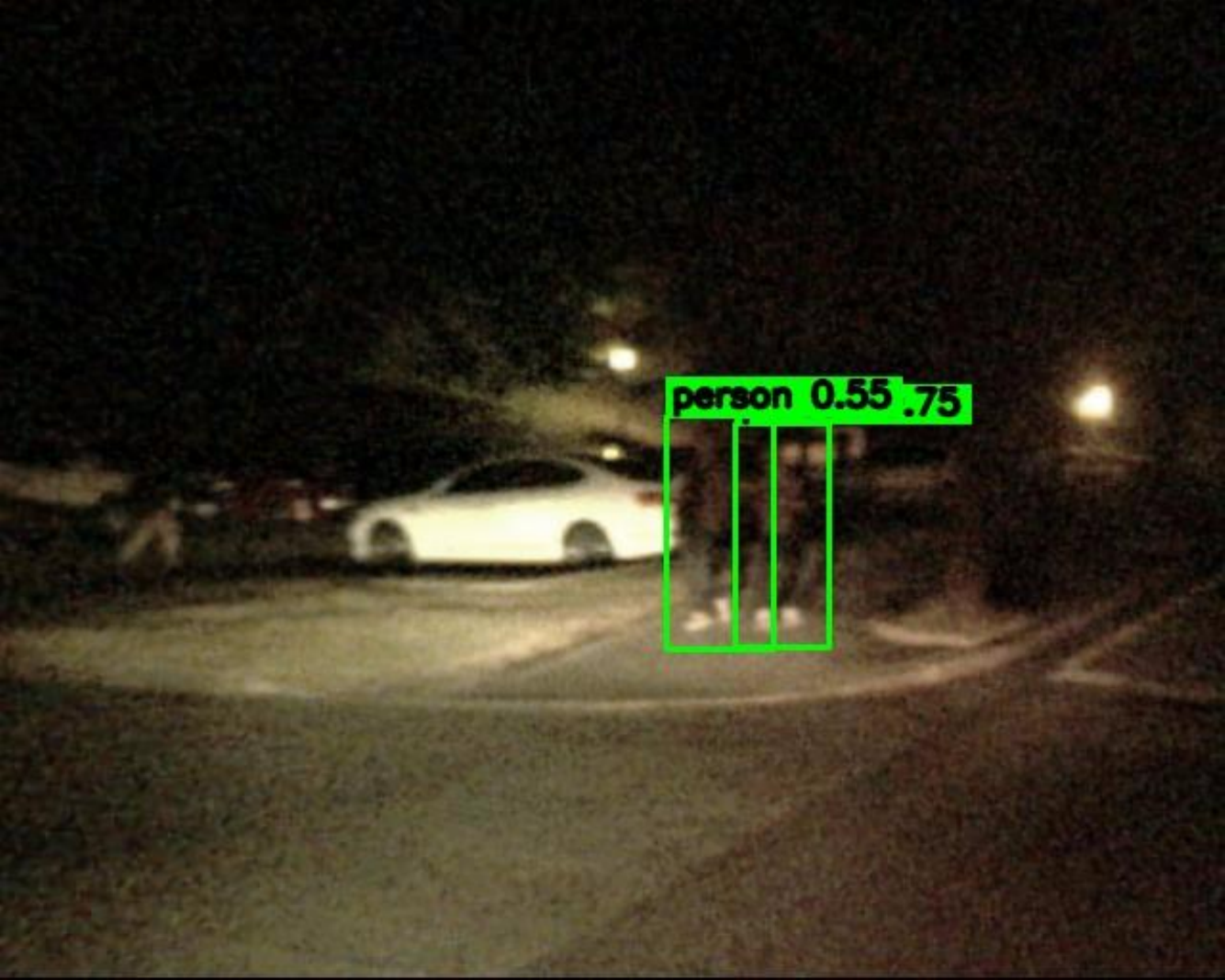} &
\includegraphics[width = 1.4in]{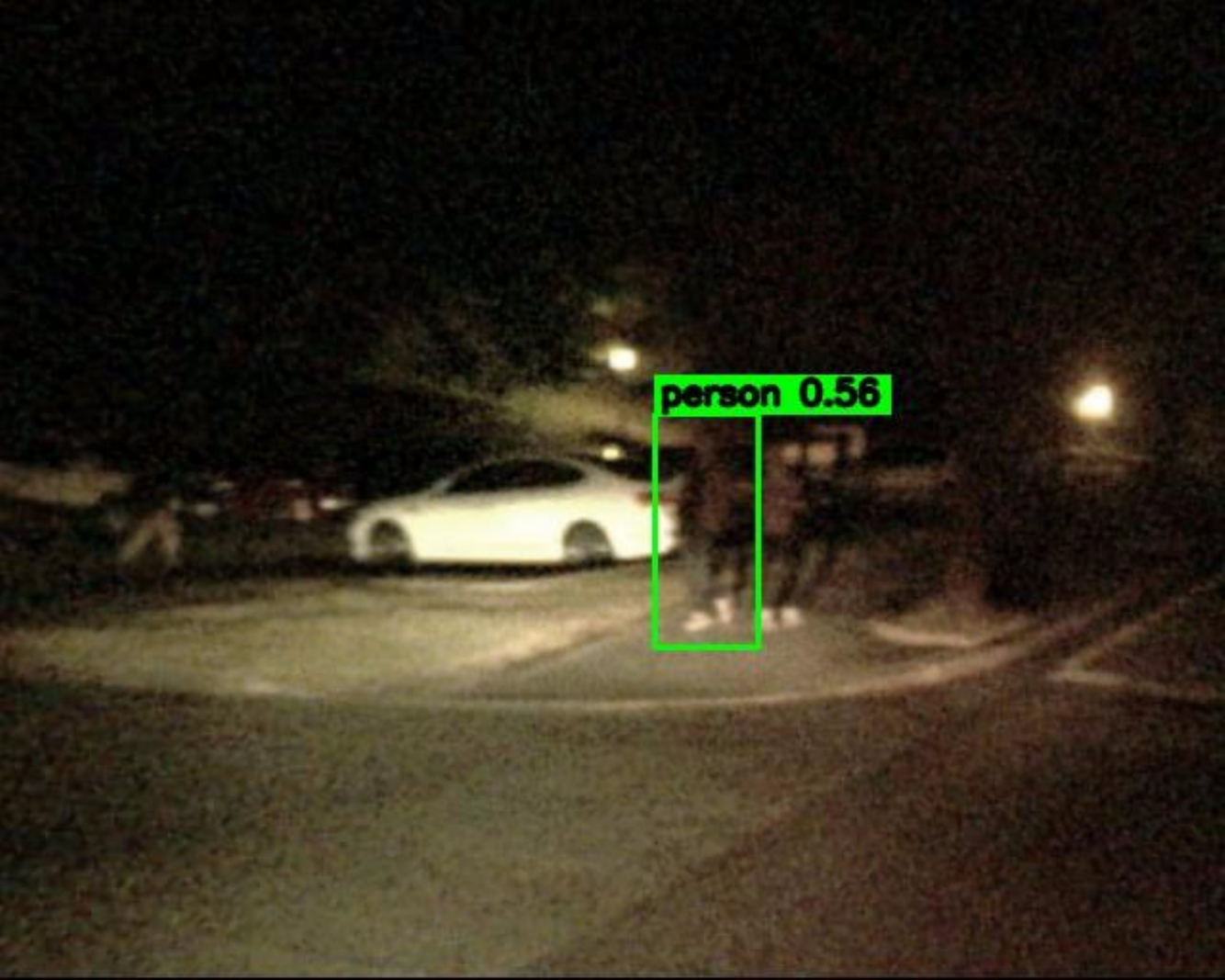} &
\includegraphics[width = 1.4in]{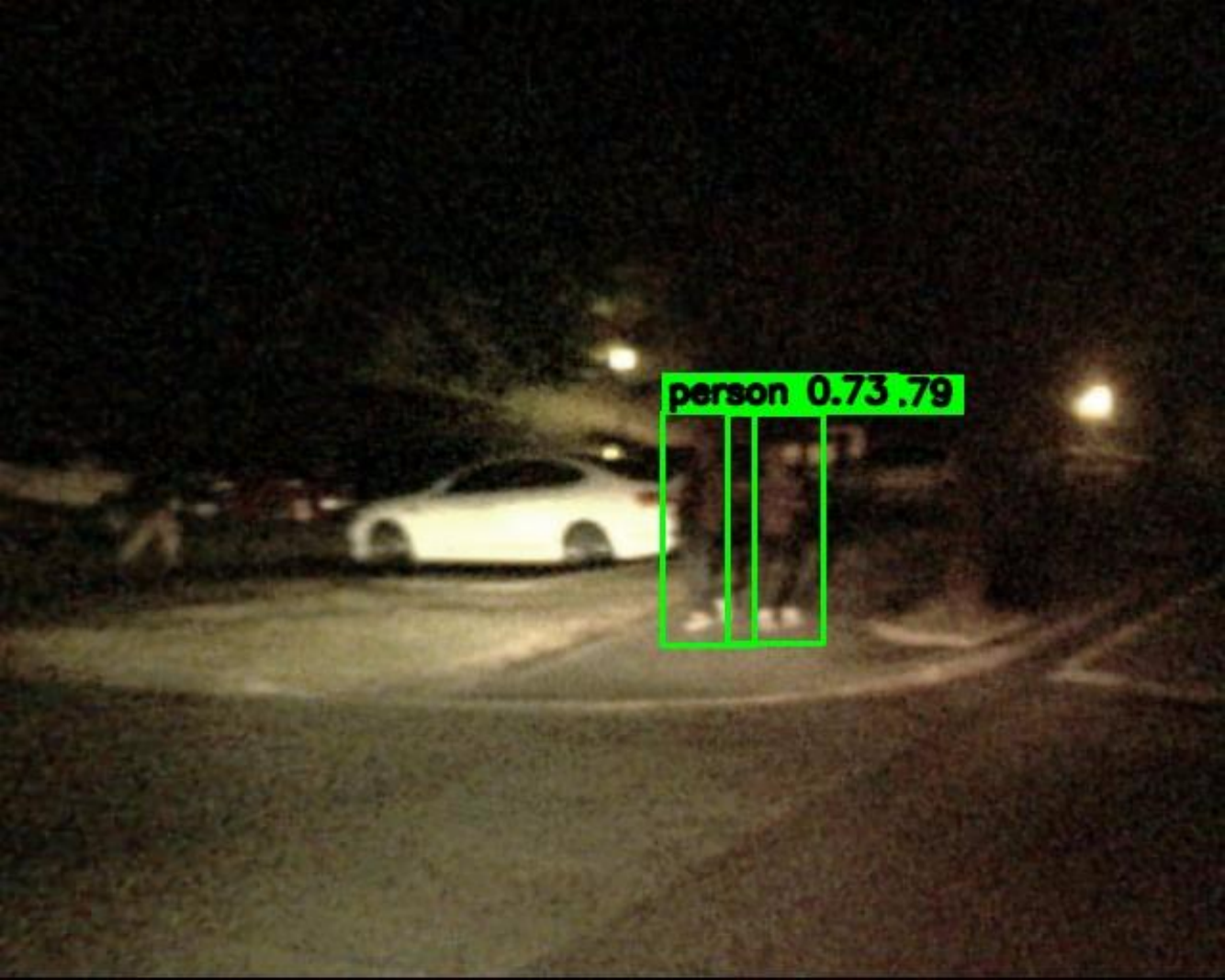} &
\includegraphics[width = 1.4in]{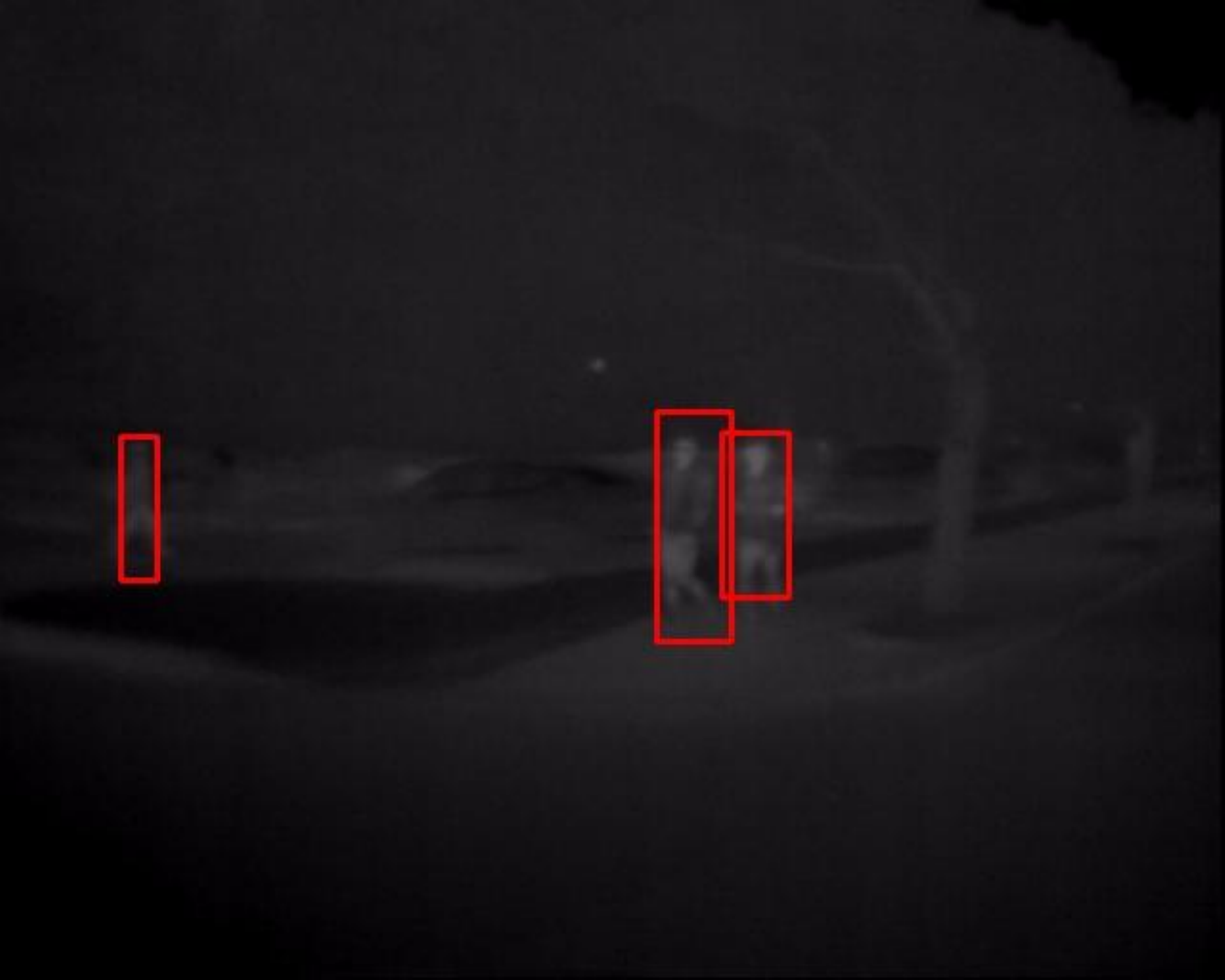} &
\includegraphics[width = 1.4in]{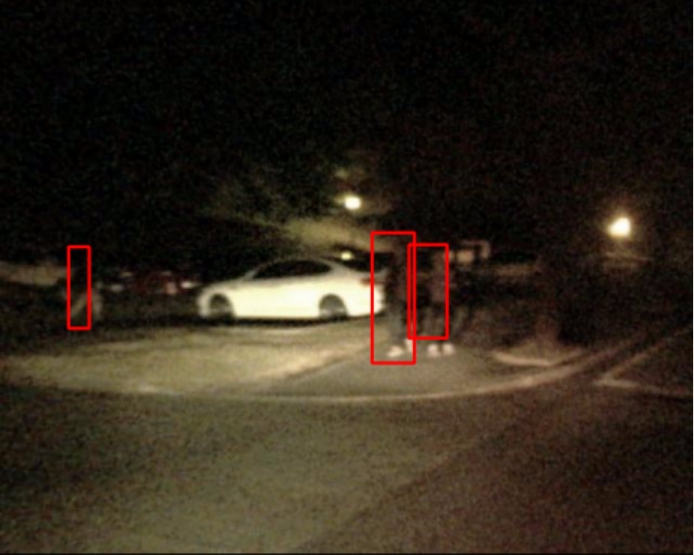} \\
\hspace{1cm} \\

& & & \bf{IR-RCNN 0.25 Conf} & \bf{IR-RCNN 0.50 Conf} & \bf{GT labels} & \bf{GT IR} & \bf{GT RGB} \\

&
&
\multirow{2}{*}[15pt]{\rotatebox{90}{\textbf{RCNN}}} &
\includegraphics[width = 1.4in]{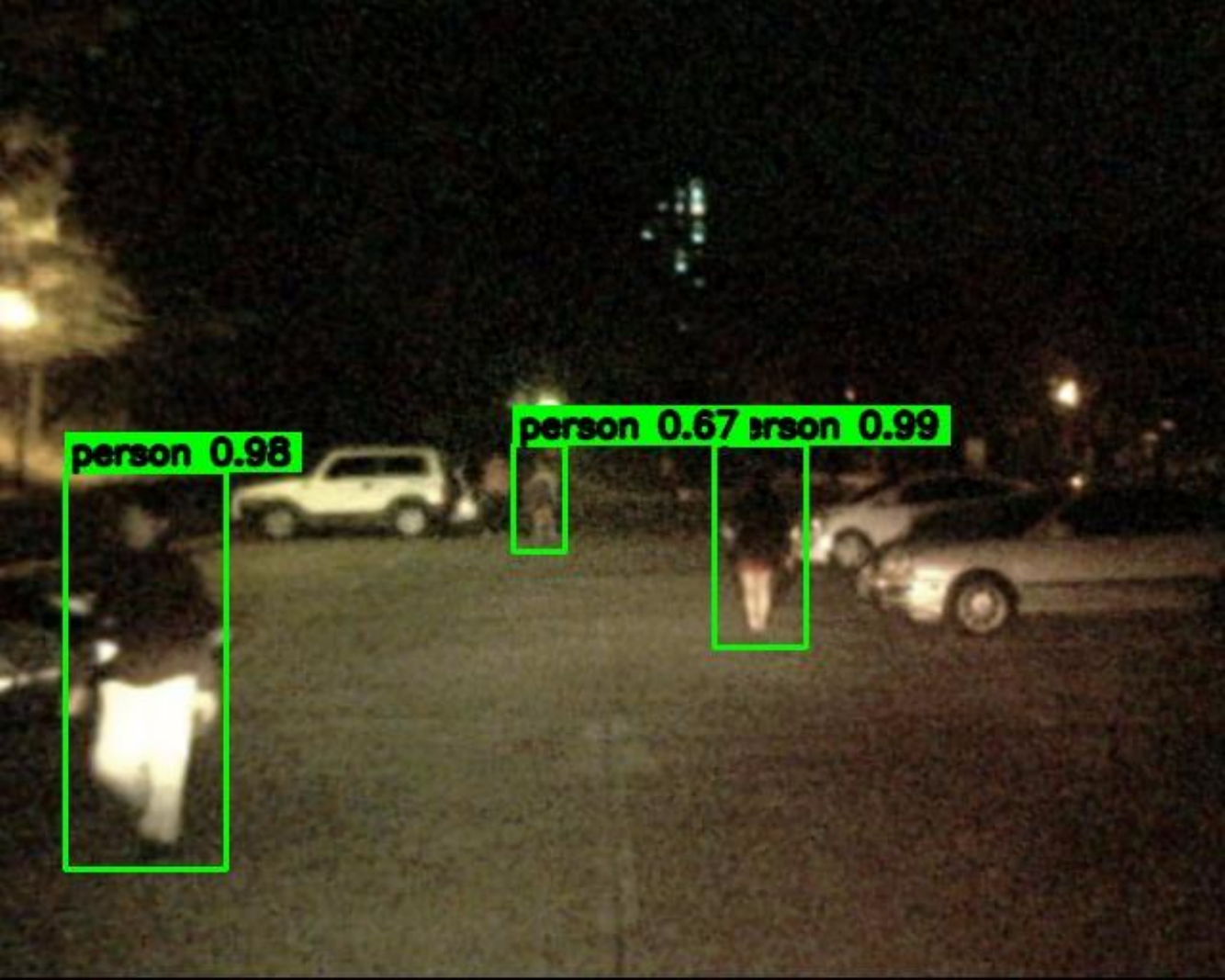} &
\includegraphics[width = 1.4in]{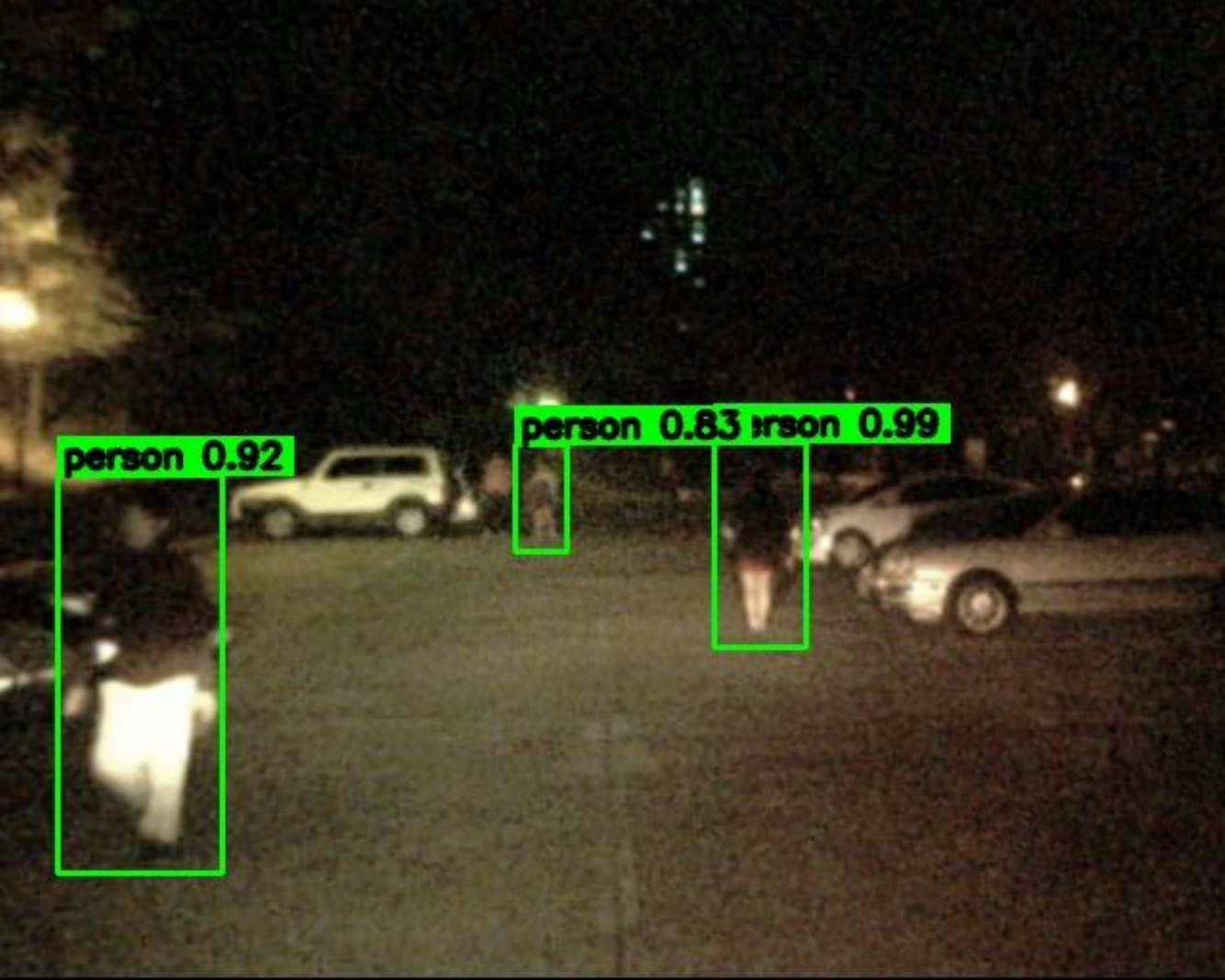} &
\includegraphics[width = 1.4in]{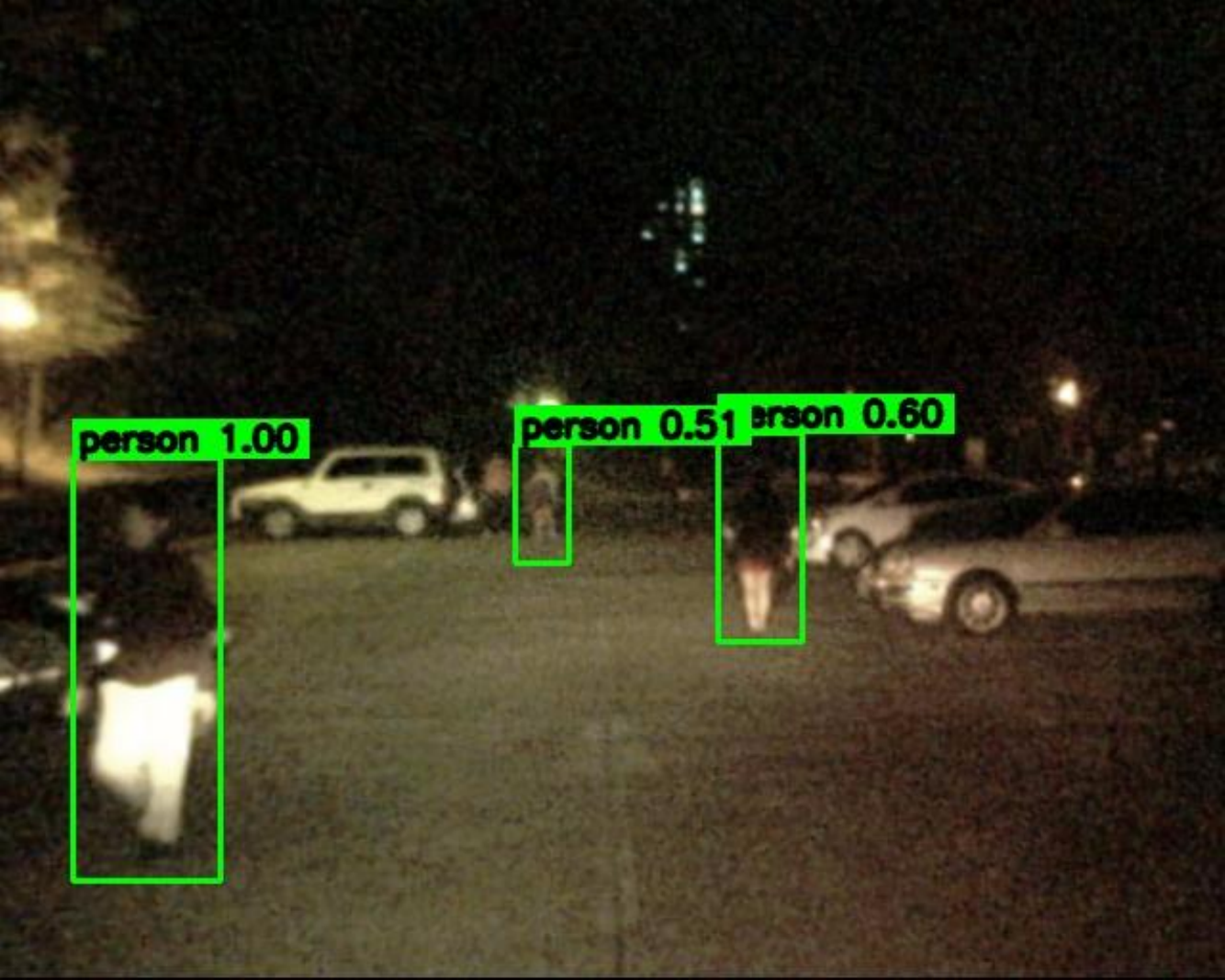} &
\includegraphics[width = 1.4in]{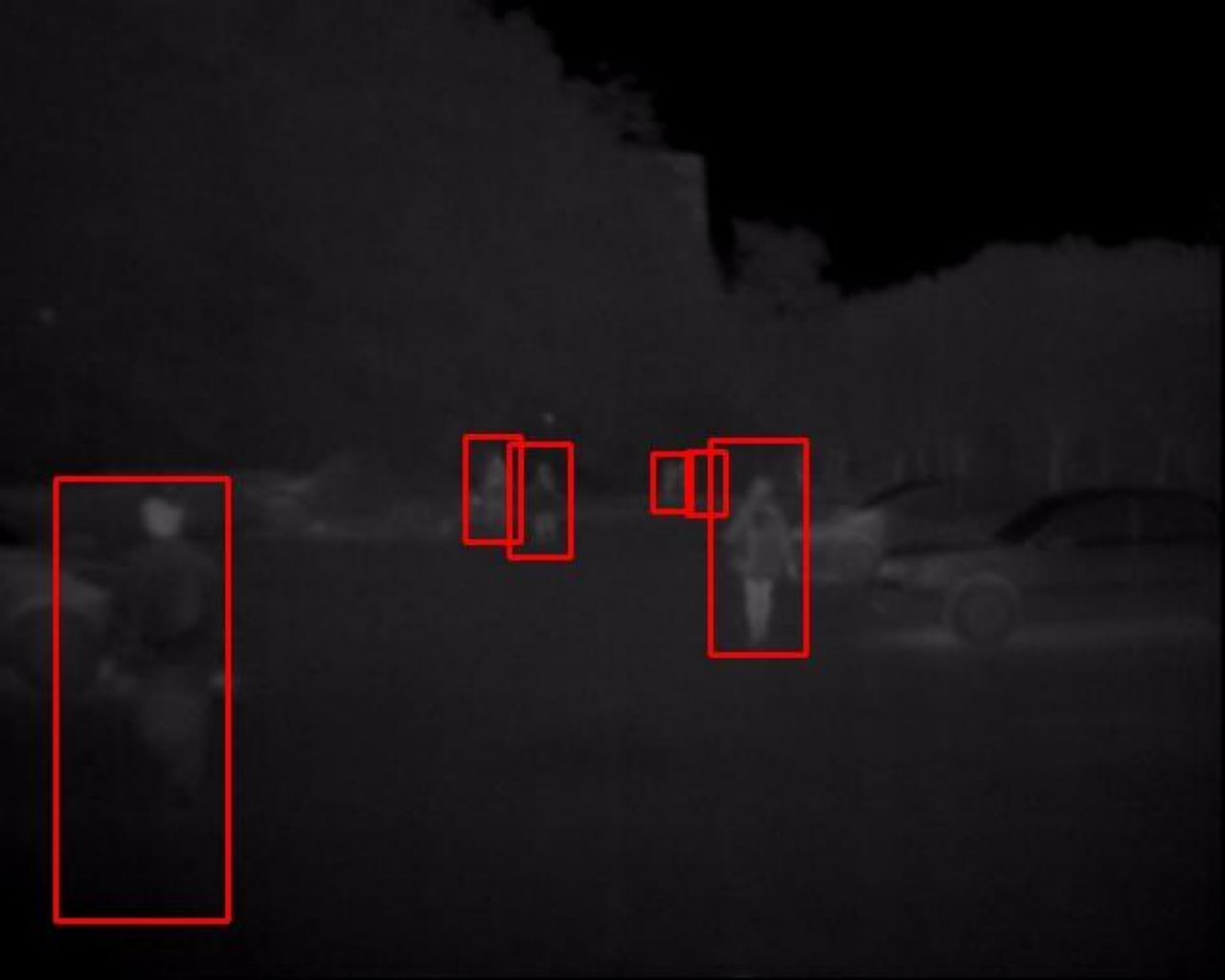} &
\includegraphics[width = 1.4in]{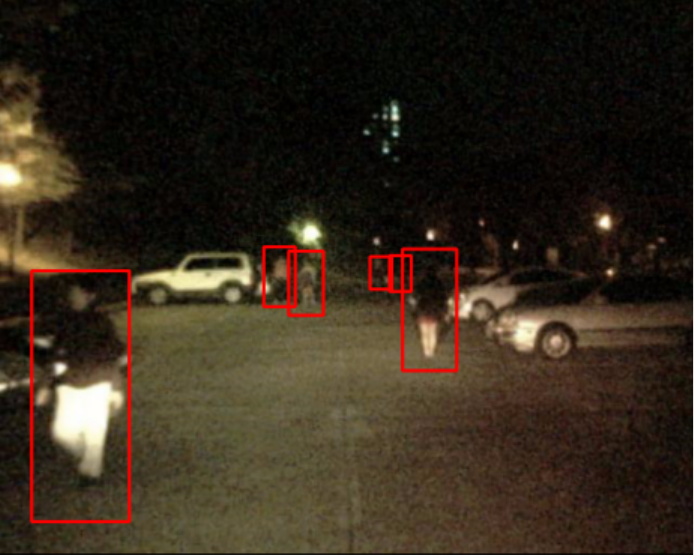}\\

&
&
&
\includegraphics[width = 1.4in]{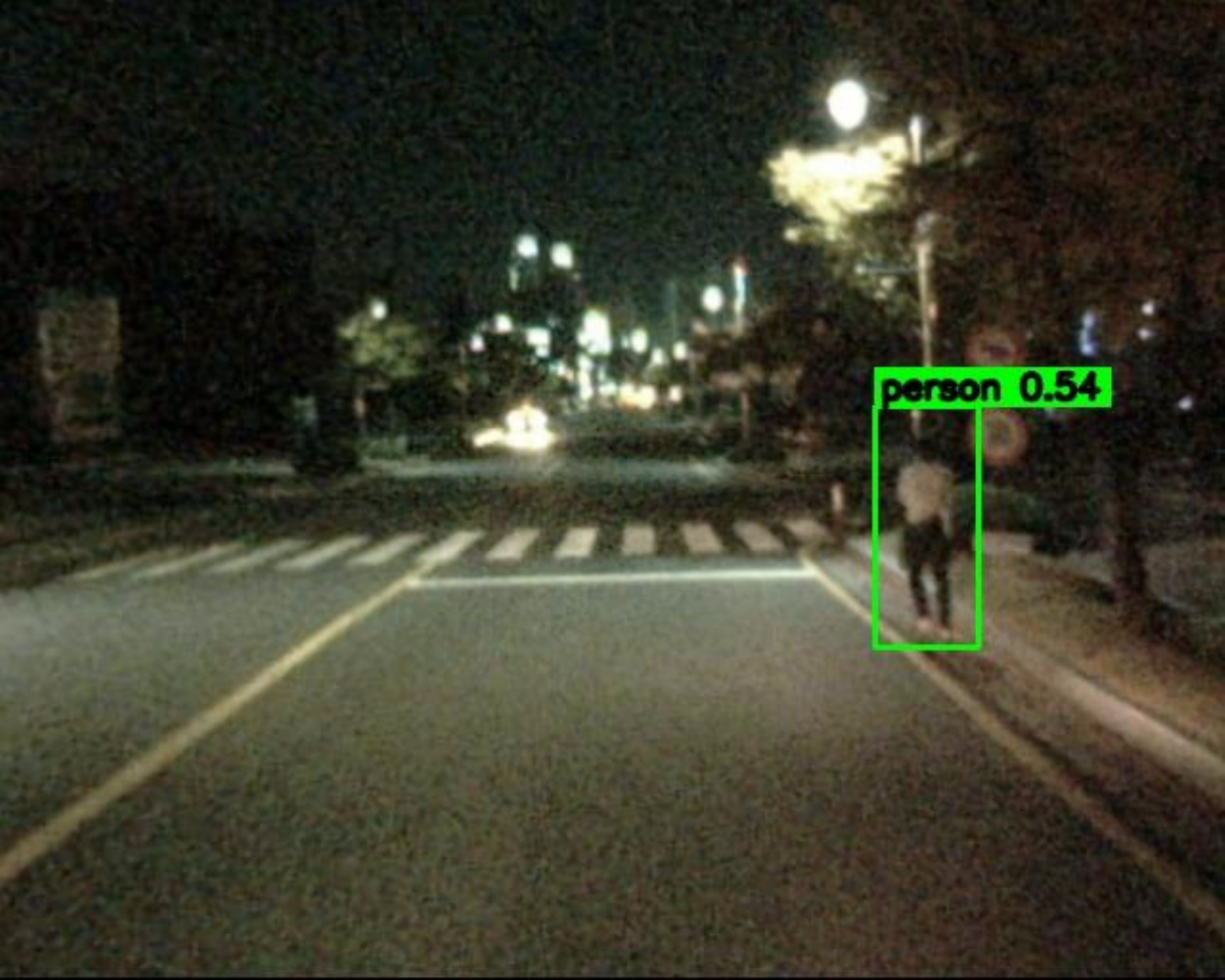} &
\includegraphics[width = 1.4in]{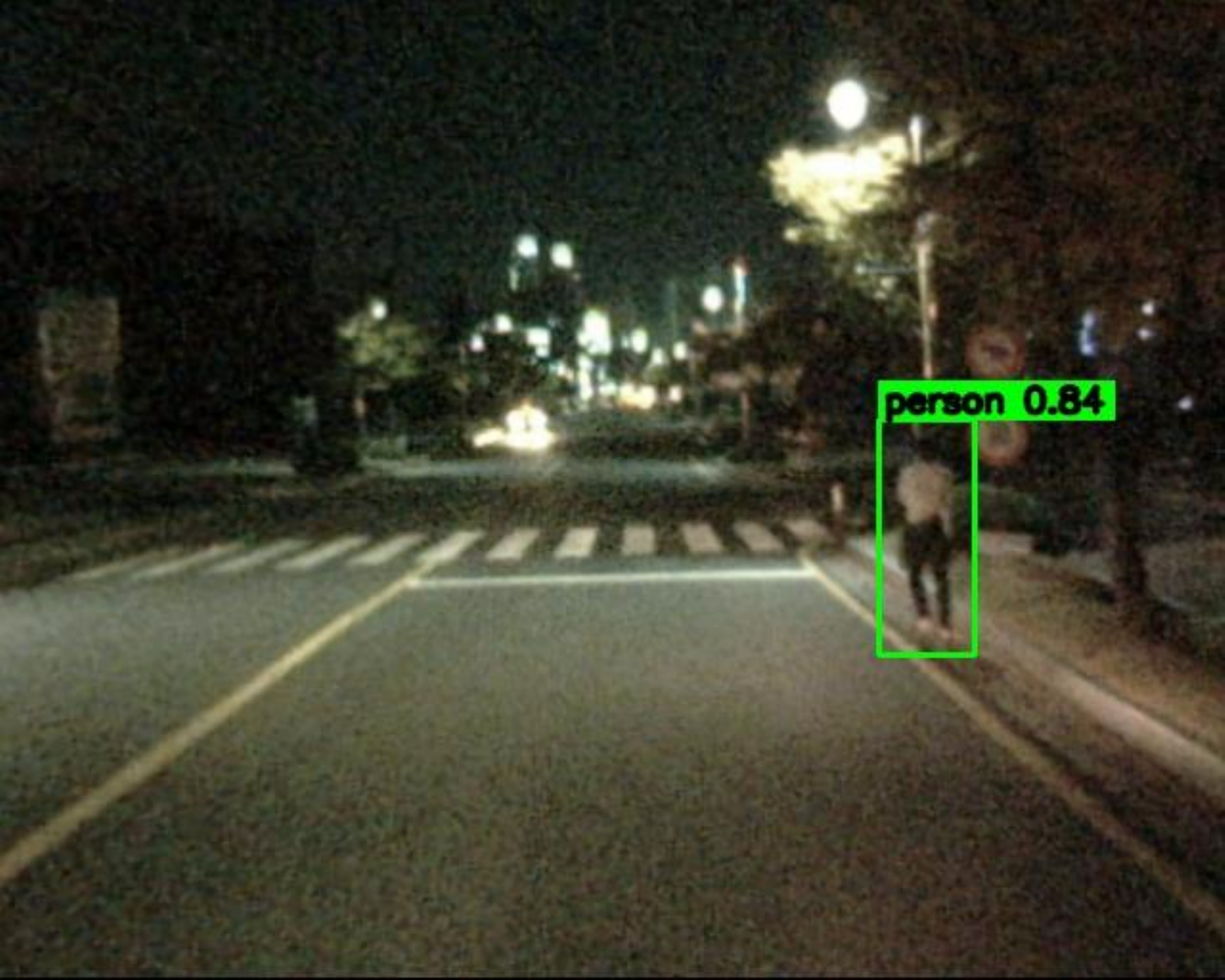} &
\includegraphics[width = 1.4in]{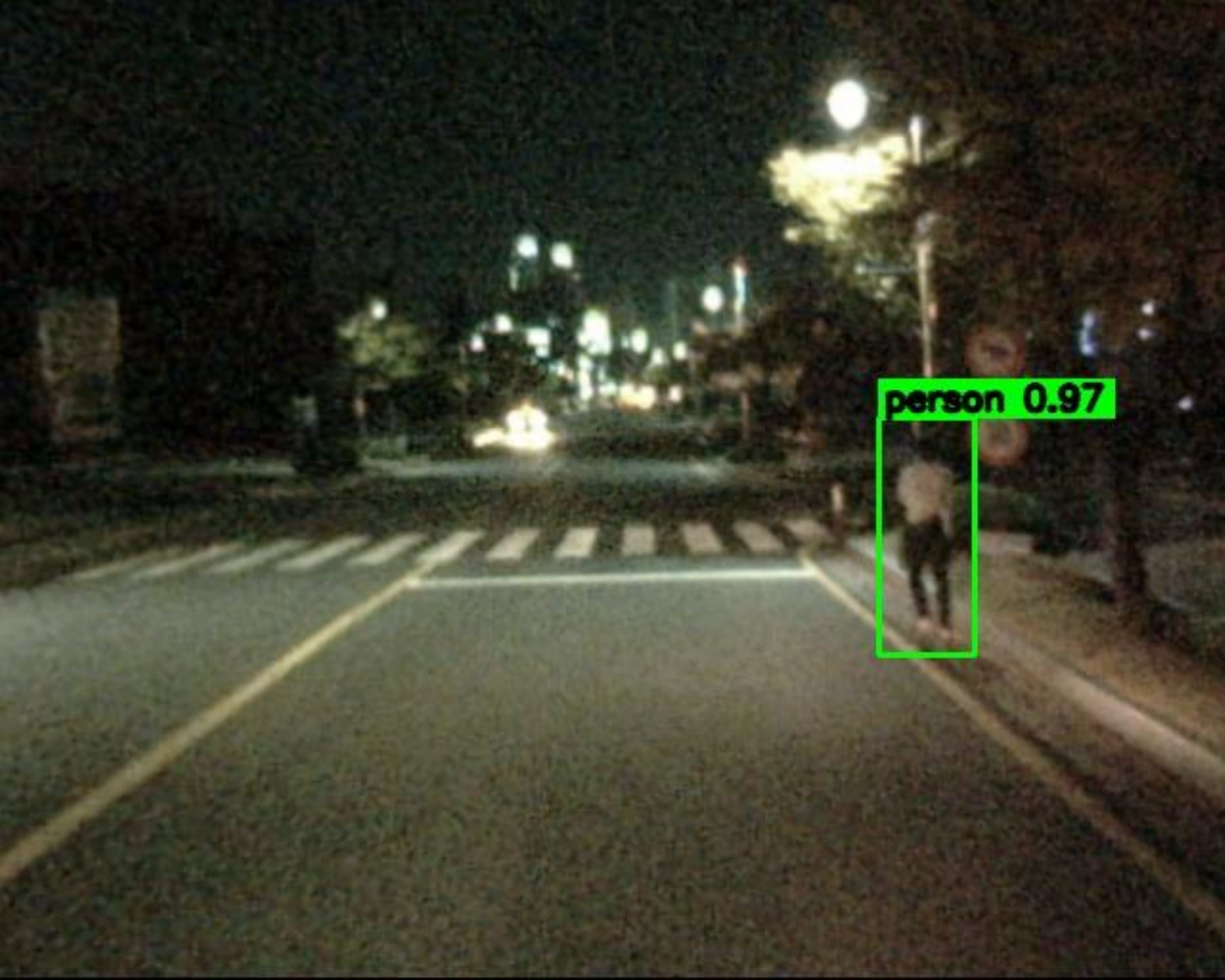} &
\includegraphics[width = 1.4in]{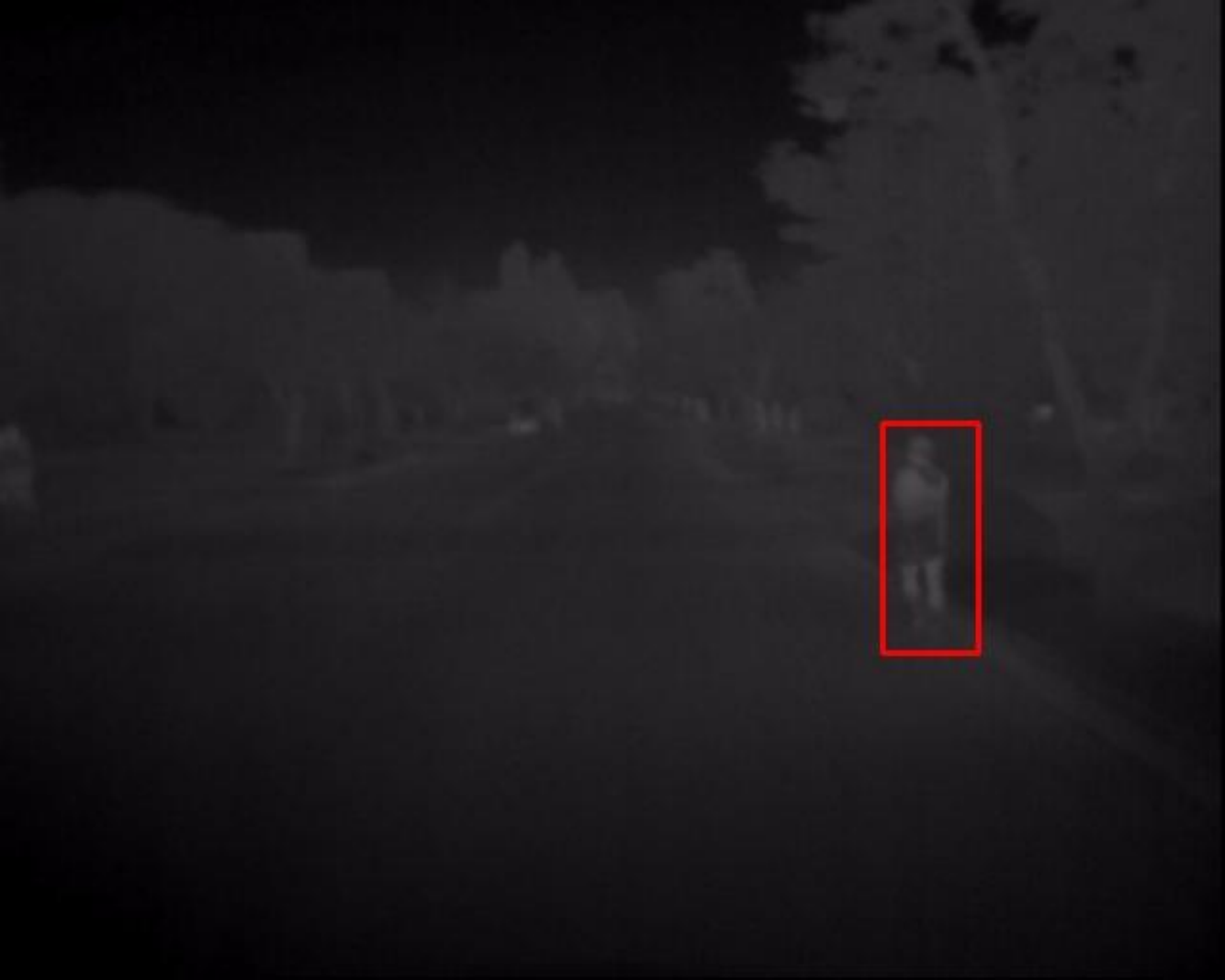} &
\includegraphics[width = 1.4in]{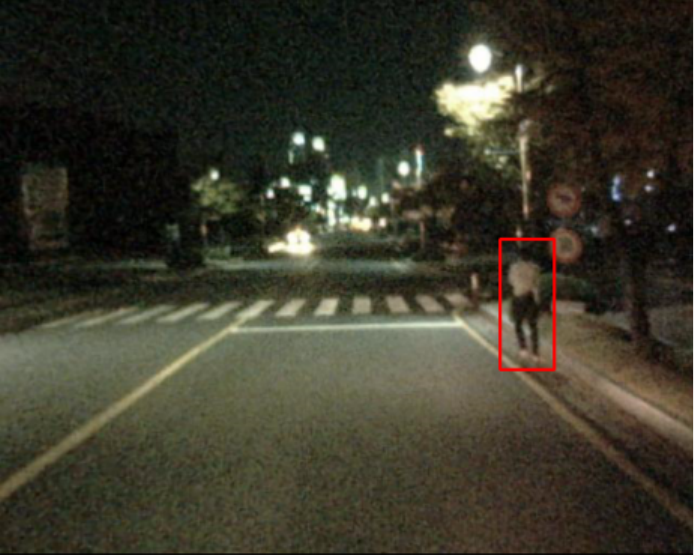} \\

\end{tabular}
} 
\caption{Examples of predictions made with different prediction model - training label combinations. Columns 1-2 correspond to the models trained with our autolabels, column 3 to the models trained with ground truth labels and columns 4 and 5 are the ground truth in infrared and RGB.  
}
\label{figgure_all_model_preds}
\end{figure*}

\section{Discussion}

This paper presented an automated labelling pipeline for low-light pedestrian detection using IR-RGB image pairs. For IR autolabelling and RGB pedestrian detection, YOLOv11l, RF-DETR, and Faster RCNN baseline models were used. The evaluation showed that training on the generated autolabels yields higher mAP@50 and lower LAMR compared to ground truth labels provided in KAIST across all tests and in 5 out of 6 for mAP@50-95, achieving the goal of matching human-generated ground truth in performance without the need for human input.

\begin{figure*}
    \centering
    \begin{tabular}{c @{\hspace{4pt}} c @{\hspace{4pt}} c @{\hspace{4pt}} c}
    \includegraphics[width=0.24\textwidth]{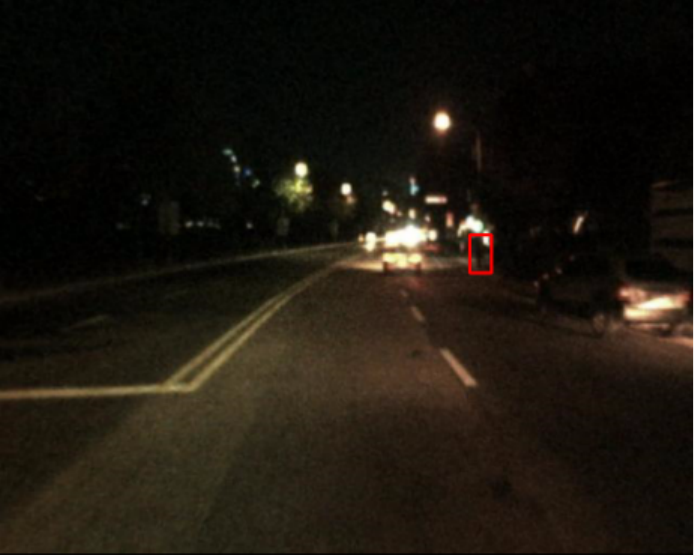} &
    \includegraphics[width=0.24\textwidth]{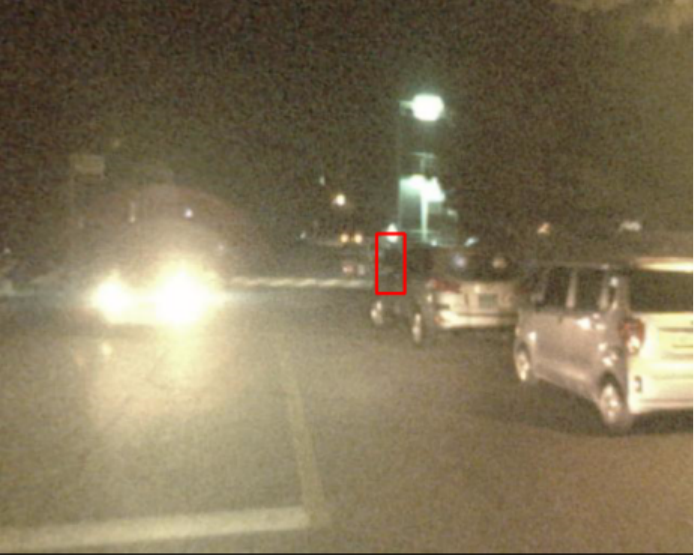} &
    \includegraphics[width=0.24\textwidth]{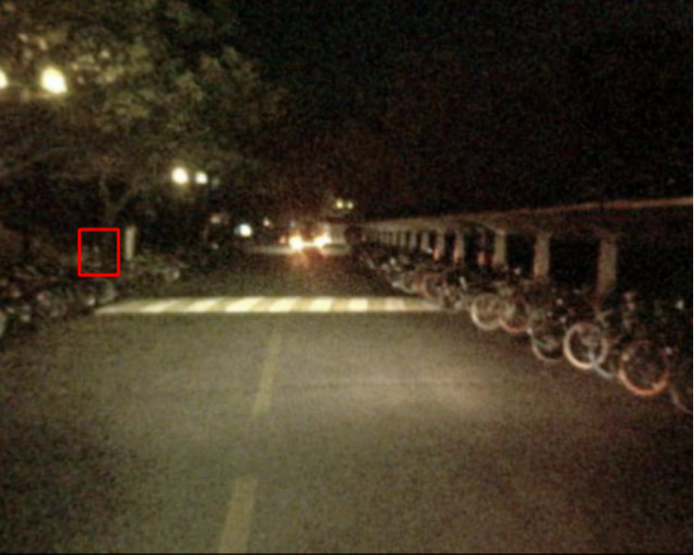} &
    \includegraphics[width=0.24\textwidth]{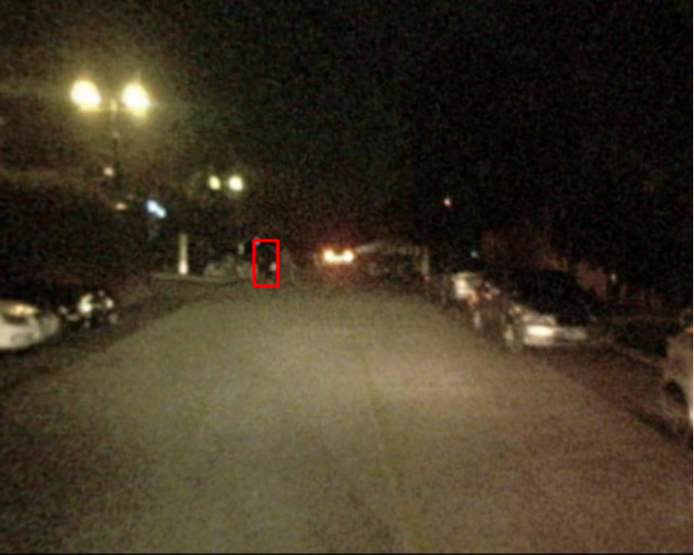}
    \end{tabular}
    \caption{KAIST ground truth difficult case labels, where pedestrians are poorly visible in the RGB frame.}
    \label{training-edgecases}
\end{figure*}

Training with the autolabels yielded better results than training with the ground truth labels, as shown in Table \ref{final_results_color_coded} and demonstrated on the testing data in Figure \ref{figgure_all_model_preds}, consequently exceeding the initial goal of matching ground truth performance. This difference in performance could be attributed to two factors: the hard cases in ground truth labels and the general difficulty of low-light pedestrian detection. In the ground truth labels, examples where pedestrians are extremely poorly visible or heavily occluded exist, with samples shown in Figure \ref{training-edgecases}. These samples can potentially act as label noise that are technically correct but lack features and visual cues, consequently lowering model performance. As object detection in low-light conditions often results in lowered accuracy, the combination of these two factors could be the reason for the performance drop. The same effect is not observed in the autolabels, as heavily occluded and partially visible pedestrians are ignored, due to the autolabelling model not detecting them as pedestrian instances, consequently filtering out extremely hard cases.   

In our experiments, another key finding was the effect of model architecture on the final output. In all cases where the three baseline models were trained on the same labels, we observe that DETR performs best, followed by YOLO and lastly by RCNN. This is observed in IR model training (Table \ref{ir_model_metrics}) and in the case where ground truth labels are used (Table \ref{final_results_color_coded}) during RGB model training. Furthermore, the effect of the model is highlighted as all DETR models performed better than any other model-label combination on the low-light RGB detection task. This observation shows the importance of appropriate model selection.

The results presented in Table \ref{final_results_color_coded} not only show the importance of quality labels and appropriate model selection, but also the challenge of low-light object detection as a whole. Of the presented models, regardless of model-label pair, all cases did not produce a model that can be reliably used as a standalone detection tool. This is concluded due to the best recall being 0.369, indicating that the models are not capable of reliably finding all pedestrians in the images. This highlights the difficulty of the RGB pedestrian detection task in low-light conditions. However, as RGB cameras are readily available in vehicles, the detection capabilities are still valuable for different intelligent systems. Larger datasets enable the training and development of more accurate models for the detection task. Our work directly contributes to this, as the proposed autolabelling method can be used for collecting extensive datasets.

Another parameter explored in this paper was the autolabelling detection confidence threshold. In all cases when lowering confidence from 0.5 to 0.25, Precision decreased, and Recall increased, F1 increased for IR-YOLO and IR-RCNN and decreased for IR-DETR (Table \ref{ir_model_metrics}). Additionally, when autolabels were generated in all cases, the models running on 0.25 confidence produced significantly more labels compared to 0.5 (Table \ref{ir_model_label_counts}). When low-light pedestrian detection models were trained (Table \ref{final_results_color_coded}), all models performed better on mAP@50, mAP@50-95, and LAMR. This shows that additional labels generated during 0.25 confidence automatic labelling had a positive effect on subsequent training, resulting in another parameter to be further explored and verifying the need for more overall data.

The proposed approach was only verified using the KAIST dataset, and further research is needed. The next steps include the gathering of a larger IR-RGB dataset to evaluate the performance consistency in larger datasets, as well as the effects of a larger number of edge case scenarios with occlusion and poor illumination. Additionally, state-of-the-art methods in infrared detection should be tested as the proposed pipeline relies solely on infrared detections and their quality. 

The proposed automatic labelling requires IR-RGB image pairs, meaning the data collection vehicle must be equipped with an IR camera and an RGB camera, which are calibrated together. Deploying such platforms can be demanding, as IR cameras are expensive and the calibration procedure can be complex. However, the IR-RGB camera setup is only needed for training data generation. Additionally, the proposed automatic labelling method enables the collection of extensive datasets with no manual human labelling. The RGB detection models trained with the autolabelled data only require an RGB camera, which is frequently available in most new cars. 

As the proposed method's results showed potential for low-light applications, similar approaches can be applied to other low-light tasks. One such task is pose estimation, as it could be implemented in a similar manner to the proposed method. Such research could aid the development of more complex systems, such as pedestrian crossing intention prediction, enabling more advanced decision-making in low-light conditions.

\bibliographystyle{apalike}
{\small
\bibliography{bibliography}}

@online{who2023traffic,
  author    = {{World Health Organization}},
  title     = {Road Traffic Injuries},
  year      = {2023},
  url       = {https://www.who.int/news-room/fact-sheets/detail/road-traffic-injuries},
  month        = {Dec},
  note      = {Accessed: 2025-04-06},
  organization = {World Health Organization}
}

@online{moore2024nightdriving,
  author       = {Milnes, Doug},
  title     = {Driving at Night Is 9 Times Deadlier Than Driving During the Day},
  year      = {2024},
  url       = {https://www.moneygeek.com/living/driving/dangerous-night-driving/},
  note      = {Accessed: 2025-04-06},
  organization = {MoneyGeek}
}

@article{yolov11_ultralytics_paper,
  author       = {Khanam, Rahima and Hussain, Muhammad},
  title        = {YOLOv11: An Overview of the Key Architectural Enhancements},
  journaltitle = {arXiv},
  year         = {2024},
  month        = {Oct},
  eprint       = {2410.17725},
  archivePrefix= {arXiv},
  primaryClass = {cs.CV},
  url          = {https://arxiv.org/abs/2410.17725}
}

@inproceedings{10YEARSOF-peddet,
  title={Ten years of pedestrian detection, what have we learned?},
  author={Benenson, Rodrigo and Omran, Mohamed and Hosang, Jan and Schiele, Bernt},
  booktitle={Computer Vision-ECCV 2014 Workshops: Zurich, Switzerland, September 6-7 and 12, 2014, Proceedings, Part II 13},
  pages={613--627},
  year={2015},
  month     = {Jan},
  doi       = {10.1007/978-3-319-16181-5_47},
  organization={Springer}
}

@article{obj_det_issues,
  title={Pedestrian detection and tracking in video surveillance system: issues, comprehensive review, and challenges},
  author={Gawande, Ujwalla and Hajari, Kamal and Golhar, Yogesh},
  journal={Recent Trends in Computational Intelligence},
  pages={1--24},
  year={2020},
  month     = {May},
  doi       = {10.5772/intechopen.90810},
  publisher={IntechOpen}
}

@ARTICLE{ped_Det,
  author={Ragesh, N. K. and Rajesh, R.},
  journal={IEEE Access}, 
  title={Pedestrian Detection in Automotive Safety: Understanding State-of-the-Art}, 
  year={2019},
  month = {Apr},
  volume={7},
  number={},
  pages={47864-47890},
  doi={10.1109/ACCESS.2019.2909992}
}

@inproceedings{domain_adaptation_autolabelling,
  title={Unsupervised domain adaptation for multispectral pedestrian detection},
  author={Guan, Dayan and Luo, Xing and Cao, Yanpeng and Yang, Jiangxin and Cao, Yanlong and Vosselman, George and Ying Yang, Michael},
  booktitle={Proceedings of the IEEE/CVF Conference on Computer Vision and Pattern Recognition Workshops},
  pages={0--0},
  year={2019},
  month     = {June},
  doi       = {10.1109/CVPRW.2019.00057},
  publisher = {IEEE}
}

@article{usupervised_ped_det_ir-thermal_autolabelling,
  title={An unsupervised transfer learning framework for visible-thermal pedestrian detection},
  author={Lyu, Chengjin and Heyer, Patrick and Goossens, Bart and Philips, Wilfried},
  journal={Sensors},
  volume={22},
  number={12},
  pages={4416},
  year={2022},
  publisher={MDPI},
  month        = {June},
  doi          = {10.3390/s22124416}
}

@InProceedings{KAIST_dataset,
author = {Hwang, Soonmin and Park, Jaesik and Kim, Namil and Choi, Yukyung and So Kweon, In},
title = {Multispectral Pedestrian Detection: Benchmark Dataset and Baseline},
booktitle = {Proceedings of the IEEE Conference on Computer Vision and Pattern Recognition (CVPR)},
month = {June},
year = {2015},
month     = {June},
doi       = {10.1109/CVPR.2015.7298706},
publisher = {IEEE}
}

@misc{FLIRADAS_dataset,
  title        = {FLIR Thermal Dataset for ADAS},
  author       = {{FLIR Systems, Inc.}},
  year         = {2022},
  month        = {January},
  url          = {https://oem.flir.com/solutions/automotive/adas-dataset-form/},
  note         = {Accessed: 2025-10-28},
  organization = {FLIR Systems, Inc.}
}

@article{KITTI-DATASET,
  title={Vision meets robotics: The kitti dataset},
  author={Geiger, Andreas and Lenz, Philip and Stiller, Christoph and Urtasun, Raquel},
  journal={The international journal of robotics research},
  volume={32},
  number={11},
  pages={1231--1237},
  year={2013},
  publisher={Sage Publications Sage UK: London, England},
  month       = {Sep},
  doi         = {10.1177/0278364913491297},
  publisher   = {SAGE Publications}
}

@article{widerperson_dataset,
  title={Widerperson: A diverse dataset for dense pedestrian detection in the wild},
  author={Zhang, Shifeng and Xie, Yiliang and Wan, Jun and Xia, Hansheng and Li, Stan Z and Guo, Guodong},
  journal={IEEE Transactions on Multimedia},
  volume={22},
  number={2},
  pages={380--393},
  year={2020},
  month       = {Feb},
  doi         = {10.1109/TMM.2019.2929005},
  publisher={IEEE}
}

@inproceedings{nightowls_dataset,
  title={Nightowls: A pedestrians at night dataset},
  author={Neumann, Luk{\'a}{\v{s}} and Karg, Michelle and Zhang, Shanshan and Scharfenberger, Christian and Piegert, Eric and Mistr, Sarah and Prokofyeva, Olga and Thiel, Robert and Vedaldi, Andrea and Zisserman, Andrew and others},
  booktitle={Computer Vision--ACCV 2018: 14th Asian Conference on Computer Vision, Perth, Australia, December 2--6, 2018, Revised Selected Papers, Part I 14},
  pages={691--705},
  year={2019},
  doi = {10.1007/978-3-030-20887-5_43},
  organization={Springer}
}

@InProceedings{CityPersons-DATASET,
author = {Zhang, Shanshan and Benenson, Rodrigo and Schiele, Bernt},
title = {CityPersons: A Diverse Dataset for Pedestrian Detection},
booktitle = {Proceedings of the IEEE Conference on Computer Vision and Pattern Recognition (CVPR)},
month = {July},
year = {2017},
doi = {10.1109/CVPR.2017.474}
}

@inproceedings{LLVIP_dataset,
  title={LLVIP: A visible-infrared paired dataset for low-light vision},
  author={Jia, Xinyu and Zhu, Chuang and Li, Minzhen and Tang, Wenqi and Zhou, Wenli},
  booktitle={Proceedings of the IEEE/CVF international conference on computer vision},
  pages={3496--3504},
  year={2021},
  url       = {https://openaccess.thecvf.com/content/ICCV2021W/RLQ/papers/Jia_LLVIP_A_Visible-Infrared_Paired_Dataset_for_Low-Light_Vision_ICCVW_2021_paper.pdf}

}

@article{cvc14_dataset,
  author       = {Gonz{\'a}lez, Alejandro and Fang, Zhijie and Socarr{\'a}s, Yainuvis and Serrat, Joan and V{\'a}zquez, David and Xu, Jiaolong and L{\'o}pez, Antonio M.},
  title        = {Pedestrian Detection at Day/Night Time with Visible and FIR Cameras: A Comparison},
  journaltitle = {Sensors},
  volume       = {16},
  number       = {6},
  pages        = {820},
  year         = {2016},
  month        = {June},
  doi          = {10.3390/s16060820}
}

@inproceedings{COCO_dataset,
  author    = {Lin, Tsung{-}Yi and Maire, Michael and Belongie, Serge and Hays, James and Perona, Pietro and Ramanan, Deva and Doll{\'a}r, Piotr and Zitnick, C. Lawrence},
  title     = {Microsoft {COCO}: Common Objects in Context},
  booktitle = {Computer Vision -- ECCV 2014, Proceedings Part V},
  pages     = {740--755},
  year      = {2014},
  month     = {September},
  doi       = {10.1007/978-3-319-10602-1_48}
}

@software{RF-DETR_2025_citation,
  author = {Robinson, Isaac and Robicheaux, Peter and Popov, Matvei},
  title  = {{RF-DETR}},
  year   = {2025},
  month  = {March},
  url    = {https://github.com/roboflow/rf-detr},
  license= {Apache-2.0}
}

@article{rcnn_citation,
  author       = {Ren, Shaoqing and He, Kaiming and Girshick, Ross and Sun, Jian},
  title        = {Faster {R-CNN}: Towards Real-Time Object Detection with Region Proposal Networks},
  journaltitle = {Advances in Neural Information Processing Systems},
  volume       = {28},
  pages        = {91--99},
  year         = {2015},
  month        = {December}
}

@inproceedings{redmon2016you,
  title={You only look once: Unified, real-time object detection},
  author={Redmon, Joseph and Divvala, Santosh and Girshick, Ross and Farhadi, Ali},
  booktitle={Proceedings of the IEEE conference on computer vision and pattern recognition},
  pages={779--788},
  year={2016}
}

@inproceedings{carion2020end,
  title={End-to-end object detection with transformers},
  author={Carion, Nicolas and Massa, Francisco and Synnaeve, Gabriel and Usunier, Nicolas and Kirillov, Alexander and Zagoruyko, Sergey},
  booktitle={European conference on computer vision},
  pages={213--229},
  year={2020},
  organization={Springer}
}

@inproceedings{girshick2015fast,
  title={Fast r-cnn},
  author={Girshick, Ross},
  booktitle={Proceedings of the IEEE international conference on computer vision},
  pages={1440--1448},
  year={2015}
}

@inproceedings{girshick2014rich,
  title={Rich feature hierarchies for accurate object detection and semantic segmentation},
  author={Girshick, Ross and Donahue, Jeff and Darrell, Trevor and Malik, Jitendra},
  booktitle={Proceedings of the IEEE conference on computer vision and pattern recognition},
  pages={580--587},
  year={2014}
}

@inproceedings{dollar2009pedestrian,
  title={Pedestrian detection: A benchmark},
  author={Doll{\'a}r, Piotr and Wojek, Christian and Schiele, Bernt and Perona, Pietro},
  booktitle={2009 IEEE conference on computer vision and pattern recognition},
  pages={304--311},
  year={2009},
  organization={IEEE}
}

@inproceedings{cai2023retinexformer,
  title={Retinexformer: One-stage retinex-based transformer for low-light image enhancement},
  author={Cai, Yuanhao and Bian, Hao and Lin, Jing and Wang, Haoqian and Timofte, Radu and Zhang, Yulun},
  booktitle={Proceedings of the IEEE/CVF international conference on computer vision},
  pages={12504--12513},
  year={2023}
}

@article{wei2018deep,
  title={Deep retinex decomposition for low-light enhancement},
  author={Wei, Chen and Wang, Wenjing and Yang, Wenhan and Liu, Jiaying},
  journal={arXiv preprint arXiv:1808.04560},
  year={2018}
}

@inproceedings{guo2021dynamic,
  title={Dynamic low-light image enhancement for object detection via end-to-end training},
  author={Guo, Haifeng and Lu, Tong and Wu, Yirui},
  booktitle={2020 25th International Conference on Pattern Recognition (ICPR)},
  pages={5611--5618},
  year={2021},
  organization={IEEE}
}

@inproceedings{liu2022image,
  title={Image-adaptive YOLO for object detection in adverse weather conditions},
  author={Liu, Wenyu and Ren, Gaofeng and Yu, Runsheng and Guo, Shi and Zhu, Jianke and Zhang, Lei},
  booktitle={Proceedings of the AAAI conference on artificial intelligence},
  volume={36},
  number={2},
  pages={1792--1800},
  year={2022}
}

@inproceedings{bai2022transfusion,
  title={Transfusion: Robust lidar-camera fusion for 3d object detection with transformers},
  author={Bai, Xuyang and Hu, Zeyu and Zhu, Xinge and Huang, Qingqiu and Chen, Yilun and Fu, Hongbo and Tai, Chiew-Lan},
  booktitle={Proceedings of the IEEE/CVF conference on computer vision and pattern recognition},
  pages={1090--1099},
  year={2022}
}

@article{zhang2022robust,
  title={Robust-FusionNet: Deep multimodal sensor fusion for 3-D object detection under severe weather conditions},
  author={Zhang, Cheng and Wang, Hai and Cai, Yingfeng and Chen, Long and Li, Yicheng and Sotelo, Miguel Angel and Li, Zhixiong},
  journal={IEEE Transactions on Instrumentation and Measurement},
  volume={71},
  pages={1--13},
  year={2022},
  publisher={IEEE}
}

@article{zhang2024tfdet,
  title={TFDet: Target-aware fusion for RGB-T pedestrian detection},
  author={Zhang, Xue and Zhang, Xiaohan and Wang, Jiangtao and Ying, Jiacheng and Sheng, Zehua and Yu, Heng and Li, Chunguang and Shen, Hui-Liang},
  journal={IEEE Transactions on Neural Networks and Learning Systems},
  year={2024},
  publisher={IEEE}
}

@inproceedings{kruthiventi2017low,
  title={Low-light pedestrian detection from RGB images using multi-modal knowledge distillation},
  author={Kruthiventi, Srinivas SS and Sahay, Pratyush and Biswal, Rajesh},
  booktitle={2017 IEEE International Conference on Image Processing (ICIP)},
  pages={4207--4211},
  year={2017},
  organization={IEEE}
}

@article{liu2021deep,
  title={Deep cross-modal representation learning and distillation for illumination-invariant pedestrian detection},
  author={Liu, Tianshan and Lam, Kin-Man and Zhao, Rui and Qiu, Guoping},
  journal={IEEE Transactions on Circuits and Systems for Video Technology},
  volume={32},
  number={1},
  pages={315--329},
  year={2021},
  publisher={IEEE}
}

@inproceedings{canton2024automatic,
  title={Automatic Labeling for Thermal Imaging Datasets Generation},
  author={Cant{\'o}n, Daniel and L{\'a}zaro, Mar{\'\i}a T},
  booktitle={2024 7th Iberian Robotics Conference (ROBOT)},
  pages={1--7},
  year={2024},
  organization={IEEE}
}

\end{document}